\documentclass{article}

\PassOptionsToPackage{numbers, compress, sort}{natbib}

\usepackage[main, final]{neurips_2026}

\usepackage[utf8]{inputenc} %
\usepackage[T1]{fontenc}    %
\usepackage{hyperref}       %
\usepackage{url}            %
\usepackage{booktabs}       %
\usepackage{amsfonts}       %
\usepackage{nicefrac}       %
\usepackage{microtype}      %
\usepackage{xcolor}         %

\usepackage{graphicx}
\usepackage{subcaption}
\usepackage{amsmath}
\usepackage{amssymb}
\usepackage{mathtools}
\usepackage{amsthm}
\usepackage{caption}
\usepackage{multirow}
\usepackage{float}
\usepackage{makecell}
\usepackage[capitalize,noabbrev]{cleveref}
\usepackage{tikz}
\usepackage[absolute]{textpos}

\theoremstyle{plain}
\newtheorem{theorem}{Theorem}[section]

\theoremstyle{definition}
\newtheorem{definition}[theorem]{Definition}

\theoremstyle{remark}

\definecolor{increase}{HTML}{0D8D0D}

\definecolor{enumcolor}{HTML}{BD282A}
\newcommand{\circlednum}[2][enumcolor]{\strut\raisebox{-1pt}{\tikz{\node[circle,inner sep=0.6pt,fill=#1,font=\scriptsize\bfseries\color{white}]{#2};}}}

\title{Parabolic Positional Encoding: A Vision-Centric, Principled, Extrapolatable, and General Approach}

\author{%
    Christoffer Koo Øhrstrøm$^{1,2}$ \and Rafael I. Cabral Muchacho$^3$ \qquad Yifei Dong$^3$
    \and Filippos Moumtzidellis$^{1,2}$ \qquad Ronja Güldenring$^{1,2}$ \qquad Florian T. Pokorny$^3$ \qquad Lazaros Nalpantidis$^{1,2}$
    \and
    \\$^1$ Technical University of Denmark
    \\Denmark
    \and
    \\$^2$ Pioneer Centre for Artificial Intelligence
    \\Denmark
    \and
    $^3$ KTH Royal Institute of Technology
    \\Sweden
}

\begin{document}

\definecolor{somegray}{rgb}{0.5, 0.5, 0.5}
\newcommand{\darkgrayed}[1]{\textcolor{somegray}{#1}}
\begin{textblock}{10}(3, 0.5)
\begin{center}
\darkgrayed{This paper has been accepted for publication at \\
Neural Information Processing Systems (NeurIPS), 2026, Spotlight}
\end{center}
\end{textblock}

\maketitle

\begin{center}
    \captionsetup{type=figure}
    \includegraphics[width=1\linewidth]{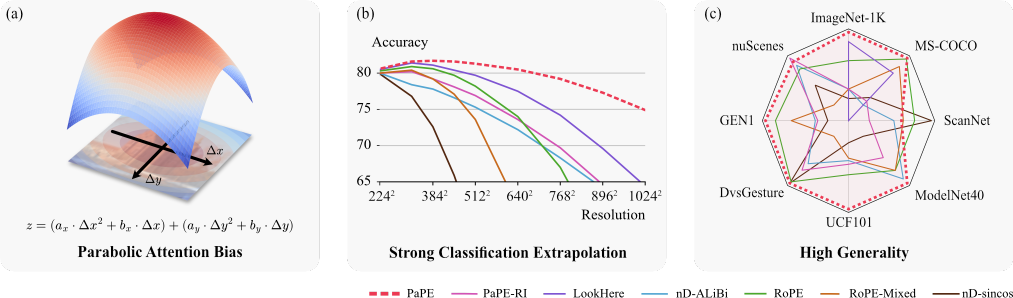}
    \captionof{figure}{
    \textbf{Parabolic Positional Encoding (PaPE).}
    \textbf{(a)} PaPE encodes positions using an attention bias based on a sum of learnable parabolas with the relative position between tokens as the dependent variable.
    \textbf{(b)} PaPE has remarkable classification extrapolation, showing high robustness beyond the training resolution of $224^2$.
    \textbf{(c)} PaPE is a general vision positional encoding---on 8 datasets that span 4 vision modalities, PaPE matches the best baseline on 5 datasets and exceeds all on 2 datasets.
    }
    \label{fig:fpf}
\end{center}

\begin{abstract}
We propose Parabolic Positional Encoding (PaPE), a parabola-based positional encoding for vision modalities in attention-based architectures.
Given a set of vision tokens---such as from videos, event camera streams, images, or point clouds---our objective is to encode their positions while accounting for the characteristics of vision modalities.
Prior works have largely extended positional encodings from 1D-sequences in language to nD-structures in vision, but only with partial account of vision characteristics.
We address this gap by designing PaPE from principles distilled from prior work: translation invariance, rotation invariance (PaPE-RI), distance decay, directionality, and context awareness.
Extrapolation experiments on ImageNet-1K show how PaPE extrapolates remarkably well, improving in absolute terms by up to 10.5 over the next-best encoding.
Generality experiments on 8 datasets across 4 modalities show that PaPE is a general vision positional encoding, as PaPE matches the best baseline on 5 datasets and exceeds all on 2 datasets.
The code is available at \href{https://github.com/DTU-PAS/parabolic-position-encoding}{https://github.com/DTU-PAS/parabolic-position-encoding} and weights are available at \cite{ohrstrom2026weights}.
\end{abstract}

\section{Introduction \& Related Works}
\label{sec:intro}
In this work, we propose a positional encoding that is designed specifically for vision modalities.

Transformers \cite{vaswaniAttentionAllYou2017} are widely used for computer vision and robotics tasks.
They have proven to be highly flexible, finding applications in several vision modalities such as videos \cite{bertasiusSpaceTimeAttentionAll2021,arnabViViTVideoVision2021}, event cameras \cite{sabaterEventTransformerSparseAware2022,gehrigRecurrentVisionTransformers2023}, images \cite{dosovitskiyImageWorth16x162020}, and point clouds \cite{wuPointTransformerV32024}.
Despite their success in the vision domain, they tend to use positional encodings that are first designed for language \cite{vaswaniAttentionAllYou2017,pressTrainShortTest2021,suRoFormerEnhancedTransformer2024} and later adapted for vision \cite{wangTranslatingMathFormula2021,fullerCROMARemoteSensing2023,schenckLearningRoPEsBetter2025}.
Although adaptations \textit{can} account for vision-specific characteristics, we hypothesize that they are incomplete in their coverage of those, because some characteristics matter little in language.
This leads us to identify relevant ideas in prior works and use these as guiding design principles for our proposed positional encoding.

\textbf{Absolute positional encodings.}
\citet{vaswaniAttentionAllYou2017} propose sinusoidal positional embeddings that are added directly to the input embeddings.
These are extended from 1D to 2D by \citet{wangTranslatingMathFormula2021} by using half of the position vector for one position dimension and the other half for the second position dimension.
The extension to arbitrary dimensions follows naturally from this.
While sinusoidal positional encoding handles multiple positional dimensions, its absolute nature lacks properties that RoPE-based and attention bias methods possess.
However, it provides the exact location of each token, which can be relevant for some tasks, as we observe in \cref{sec:generality}.

\textbf{RoPE-based methods.}
RoPE \cite{suRoFormerEnhancedTransformer2024} stands out as a state-of-the-art positional encoding.
It rotates queries and keys by a product of their position and a set of fixed frequencies, so the query-key dot product encodes the relative position, yielding \textit{translation invariance}.
One of the properties of RoPE is that attention strength diminishes as tokens move farther apart, a property we denote as \textit{distance decay}.
While vanilla RoPE operates purely in 1D, Axial RoPE \cite{chuVisionLLaMAUnifiedLLaMA2024} generalizes RoPE to arbitrary dimensionality by assigning disjoint subsets of the query and key vectors to each positional dimension and applying RoPE separately to each.
However, because these dimensions are treated independently, Axial RoPE cannot naturally capture diagonal interactions.
RoPE-Mixed \cite{heoRotaryPositionEmbedding2024}, along with related approaches \cite{ostmeierLieRELieRotational2025,yuComRoPEScalableRobust2025,schenckLearningRoPEsBetter2025}, introduce distinct learnable frequencies per positional dimension to imbue \textit{directionality} into RoPE.

\textbf{Extrapolation.}
More recently, \textit{context-aware} variants of RoPE \cite{veisiContextawareRotaryPosition2025,wangPositionalEncodingTokenAware2025} have shown that letting frequencies depend on each token improves the ability of language models to generalize to sequence lengths beyond what is seen during training.
In vision, we also want models that train cheaply at low resolutions, yet seamlessly run inference at higher ones.
Extrapolation reduces training costs and is an indicator of robustness.
Moreover, strong extrapolation suggests that a model pretrained at low resolutions can be fine-tuned more easily to high-resolution target domains.
Prior works \cite{sunLengthExtrapolatableTransformer2023,heoRotaryPositionEmbedding2024,fullerLookHereVisionTransformers2024} have shown that positional encoding greatly determines the extrapolation capabilities of a model.

\textbf{Attention biases}
modify attention scores by adding a bias term directly into the attention matrix before the softmax.
This bias depends on the relative positions of the query and key, reinforcing the importance of \textit{translation invariance}.
However, these bias terms often require materializing the full attention matrix, making them incompatible with efficient attention kernels \cite{daoFlashAttentionFastMemoryEfficient2022,daoFlashAttention2FasterAttention2024}.
ALiBi \cite{pressTrainShortTest2021}, designed for autoregressive language models, subtracts a multiple of the relative distance between the tokens.
This simple mechanism enforces both \textit{translation invariance} and \textit{distance decay} of attention scores.
2D-ALiBi \cite{fullerCROMARemoteSensing2023} generalizes this idea to images by subtracting a multiple of the 2D Euclidean distance between positions, further endowing the encoding with \textit{rotation invariance}.
LookHere \cite{fullerLookHereVisionTransformers2024} extends 2D-ALiBi with \textit{directionality} by restricting the field of view of each head to different directions.
This adds \textit{directionality}, but at the cost of losing \textit{rotation invariance}.

From the prior works, we have identified the following set of guiding design principles that we consider to be important for positional encoding of vision modalities, and elaborate on in \cref{sec:design_principles}:
\textit{translation invariance}, \textit{rotation invariance}, \textit{distance decay}, \textit{directionality}, and \textit{context awareness}.
To the best of our knowledge, no existing positional encoding is derived from all five principles.
We use the principles to propose a novel positional encoding:
\textbf{Pa}rabolic \textbf{P}ositional \textbf{E}ncoding (PaPE).
PaPE treats the relative position between two tokens as the dependent variable in a sum of parabolas and uses this to encode positions through an attention bias.
However, its formulation supports separate query-key transformations, which means PaPE does not have to materialize the full attention matrix.
In addition, we propose a rotation invariant version---PaPE-RI.
We summarize our contributions and claims as follows:

\circlednum{1} \textcolor{enumcolor}{\textbf{Parabolic Positional Encoding.}} We propose PaPE and PaPE-RI: parabola-based positional encodings for vision modalities that are designed from the compiled  principles.

\circlednum{2} \textcolor{enumcolor}{\textbf{Compatibility with Efficient Attention Kernels.}} PaPE uses query-key transformations---similar to RoPE \cite{suRoFormerEnhancedTransformer2024}---to be compatible with efficient attention kernels \cite{daoFlashAttentionFastMemoryEfficient2022,daoFlashAttention2FasterAttention2024}.

\circlednum{3} \textcolor{enumcolor}{\textbf{Strong Classification Extrapolation.}} PaPE is found to extrapolate remarkably well beyond its training resolution for classification, surpassing all baselines by at least 10.5 at resolution $1024^2$.

\circlednum{4} \textcolor{enumcolor}{\textbf{High Generality.}} We evaluate PaPE on 8 datasets across 4 vision modalities (videos, event cameras, images, and point clouds) and find that PaPE matches the best baseline in 5 datasets and exceeds all baselines in 2 datasets.

\section{Preliminaries}
\label{sec:preliminaries}
Attention \cite{vaswaniAttentionAllYou2017} for a collection of tokens $X = \{x_{1}, \cdots , x_{n} \}$ with $x_{i} \in \mathbb{R}^{d}$ is given as follows where queries, keys, and values are derived through learnable matrices $W_{q}, W_{k}, W_{v} \in \mathbb{R}^{h \times d}$, so that $q_{i} = W_{q} \cdot x_{i}$, $k_{i} = W_{k} \cdot x_{i}$, and $v_{i} = W_{v} \cdot x_{i}$:
\begin{equation}
    \label{eqn:attention}
    A \cdot V = \textrm{softmax} \left( \frac{Q \cdot K^{T}}{\sqrt{h}} \right) \cdot V .
\end{equation}
For clarity, we omit attention heads here and in the next section, noting that the extension to multi-head attention is straightforward.
We define $S = Q \cdot K^{T}$ and obtain from \cref{eqn:attention} that
\begin{equation}
    \label{eqn:attention_score}
    S_{ij} = \langle q_{i}, k_{j} \rangle,
\end{equation}
i.e., token similarity is measured as the dot products between all query-key pairs.
Since attention is permutation invariant, we require positional information to encode information about the arrangement of tokens.
To this end, we associate a position vector $r_{i} \in \mathbb{R}^{p}$ with each token, where $p$ is the position dimensionality.
For example, $p = 2$ for images (x, y) and $p = 3$ for point clouds (x, y, z).
Our goal is to leverage these position vectors to inject positional information into the dot products.

\section{Design Principles}
\label{sec:design_principles}
We now describe and motivate the design principles that we identified in prior works.
These principles are not a rigid checklist for the perfect positional encoding.
They are a set of properties that we reason---and empirically evaluate in the experiments---are valuable in practice.
We give mathematical definitions of the principles when applicable.\footnote{We assume that token positions reside in a standard $p$-dimensional Euclidean space $\mathbb{R}^{p}$.
}

\textbf{Translation invariance} matters because vision tasks usually depend on patterns defined by how parts relate to each other, not by where they are.
A cat is a cat whether it appears in the top-left or bottom-right of an image.
This calls for positional encodings that are invariant to global translations.
\begin{definition}
    \label{def:translation_invariance}
    A function $f$ is translation invariant if,
    \begin{equation}
        \label{eqn:translation_invariance}
        f(r_i + t, r_j +t) = f(r_i, r_j), \forall t, r_i, r_j \in \mathbb{R}^p.
    \end{equation}
\end{definition}

\textbf{Rotation invariance} is valuable in settings where the orientation of an object should not affect the prediction.
For instance, classifying a 3D object from a point cloud should not depend on how it is rotated in space.
However, we do not treat rotation invariance as a universal requirement: orientation often carries key information about motion and action.
We therefore consider rotation invariance as a task-dependent, special-case principle.
\begin{definition}
    \label{def:rotation_invariance}
    A function $f$ is rotation invariant if,
    \begin{align} \label{eqn:rot_invariance}
        f(R r_i, R r_j) = f(r_i, r_j),\forall r_i, r_j \in \mathbb{R}^{p}, \forall R \in \textrm{SO}(p).
    \end{align}
\end{definition}

\textbf{Distance decay} captures the intuition that nearby tokens should interact more strongly than distant ones.
Specifically, attention between two tokens should decrease as their distance increases.
This biases the positional encoding towards local interactions.

\textbf{Directionality} is the ability to modulate attention not just by how far tokens are from each other, but also by which direction.
Unlike language, which is 1D and largely indifferent to geometric direction, vision is inherently directional: above, below, left, right, and diagonal often carry distinct semantic roles.
As we move to higher-dimensional settings, e.g. 3D and spatio-temporal data, the number of possible directions amplifies this effect.
A good positional encoding for vision modalities should account for these directional cues.

\textbf{Context awareness.}
Positional encodings must be capable of emphasizing local neighborhoods while also enabling long-range interactions.
At first glance, these appear to be conflicting goals.
Context awareness reconciles the two by letting the model adapt the decay and direction strength conditioned on the content of the token.
Rather than enforcing a fixed decay and direction pattern, the model can decide when to emphasize local neighborhoods and when to focus on long-range interactions.

\section[Parabolic Positional Encoding (1)]{Parabolic Positional Encoding \circlednum{1}}
\label{sec:pape}
\begin{figure*}[t]
    \centering
    \includegraphics[width=1\linewidth]{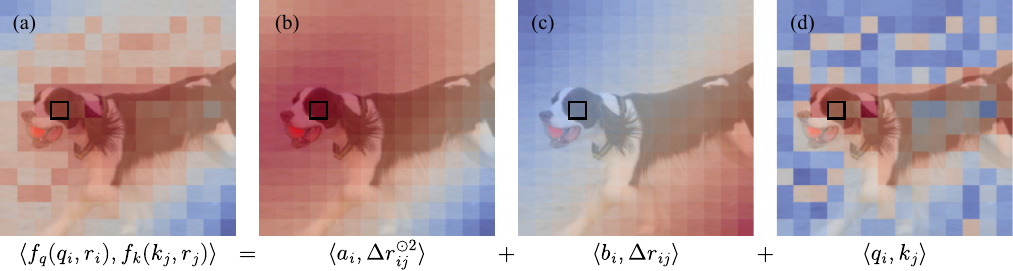}
    \caption{\textbf{Overview of Parabolic Positional Encoding (PaPE).}
    PaPE decomposes attention (a) into distance (b), direction (c), and semantics (d).
    Using the dog’s eye as the query, PaPE learns to look in a bottom-right direction, while decaying attention with distance.
    The attention (a) is compatible with efficient attention kernels through separate query-key transformations.
    Colormap: \includegraphics{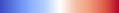}
    }
    \label{fig:method}
\end{figure*}
We now introduce \textbf{Pa}rabolic \textbf{P}ositional \textbf{E}ncoding (PaPE) and its rotation invariant version, PaPE-RI.
We begin by deriving the method and then illustrate how the design principles are incorporated into PaPE.
Afterwards, we derive query-key transformations that ensure compatibility with efficient attention kernels.

A general form 1D parabola can be written as
\begin{equation}
   y =  ax^{2} + bx + c.
\end{equation}
Our goal is to reshape the attention score in \cref{eqn:attention_score} into a sum of $m$ such parabolas. To this end, we first define
\begin{equation}
    \label{eqn:delta_pos}
    \Delta r_{ij} = W_{p} \cdot (r_{j} - r_{i}),
\end{equation}
where $W_{p} \in \mathbb{R}^{m \times p}$ is a learnable projection.\footnote{While $W_p$ is not strictly required to achieve our goal, our empirical results (see \cref{sec:ablations}) show that it is beneficial, and we motivate it in Appendix ~\ref{app:wp_m_motivation}.}
The components of $\Delta r_{ij}$ serve as the dependent variable of the parabolas.
We obtain the $a$ and $b$ coefficients directly from the token representation, $x_{i}$, using the learnable projections $W_{a}, W_{b} \in \mathbb{R}^{m \times d}$:
\begin{align}
    \label{eqn:a}
    a_{i} &= - \mathrm{softplus}\!\left( W_{a} \cdot x_{i} \right) \\
    \label{eqn:b}
    b_{i} &= W_{b} \cdot x_{i},
\end{align}
where softplus followed by negation guarantees ${a_{i} \in \mathbb{R}_{< 0}^{m}}$, making each parabola concave.
We now present the main equation that describes PaPE, where $(\cdot)^{\odot 2}$ denotes the Hadamard square:
\begin{equation}
    \label{eqn:attention_pape}
    S_{ij} = \langle a_{i}, \Delta r_{ij}^{\odot 2} \rangle + \langle b_{i}, \Delta r_{ij} \rangle + \langle q_{i}, k_{j} \rangle.
\end{equation}
\cref{fig:method} visualizes this as a sum of three attention maps.
Expanding the dot products makes the parabolic structure explicit: the attention score decomposes into a sum of $m$ general form parabolas,
\begin{equation}
    \label{eqn:attention_pape_sum}
    S_{ij} = \sum_{\ell =1}^{m} \left(
    \underbrace{a_{i\ell}  \Delta r_{ij\ell}^{2}}_{ax^{2}} +
    \underbrace{b_{i\ell}  \Delta r_{ij\ell}}_{bx} +
    \underbrace{\frac{\langle q_{i}, k_{j} \rangle}{m}}_{c} \right) .
\end{equation}
PaPE instantiates our design principles, as we will see next.

\textbf{PaPE is translation invariant}.
This follows directly from modeling the relative position, $\Delta r_{ij}$, between pairs of tokens:
$W_p ((r_j + t) - (r_i +t)) = W_p (r_j - r_i), \forall t \in \mathbb{R}^p$.

\textbf{PaPE-RI: a rotation invariant instantiation of PaPE.}
In the form given in \cref{eqn:attention_pape}, PaPE is \emph{not} rotation invariant.  
By imposing simple constraints on its parameters, we obtain a rotation invariant instantiation, which we call PaPE-RI. PaPE-RI is defined by setting all $b_i = 0$, choosing $a_{ik} = \alpha_i$ with $\alpha_i \in \mathbb{R}_{<0}$, and enforcing $W_p = w_p I_p \in \mathbb{R}^{p \times p}$ with $w_p \in \mathbb{R}$.
Under these constraints, PaPE-RI becomes provably rotation invariant; we provide the formal proof in Appendix~\ref{sec:proof-papeRI}.
An implication of setting $b_i = 0$ is that PaPE-RI loses directionality.

\textbf{PaPE has distance decay.}
Since $a_{i}$ is constrained to be negative, it follows that the square terms penalize the distances as $\Delta r_{ij}$ moves away from the origin.

\textbf{PaPE is directional.}
Recall that the dot product between two vectors $x,y$ is proportional to the cosine of the angle $\theta$ between them:
\begin{equation}
    \label{eqn:dot-product-similarity}
    \langle x, y \rangle = \|x\| \|y\| \cos (\theta).
\end{equation}
The dot product is maximized when $\theta = 0$ since ${\cos(0) = 1}$.
Therefore, $\langle b_{i}, \Delta r_{ij} \rangle$ acts as a direction term and it is largest when $\Delta r_{ij}$ is aligned with the direction defined by $b_{i}$.

\textbf{PaPE has context awareness.}
The distance decay and direction are governed by $a_{i}$ and $b_{i}$, respectively.
Because both vectors are computed from the token content itself, \cref{eqn:a,eqn:b}, PaPE naturally adapts to context instead of relying on a fixed pattern.
This enables PaPE to support long-range interactions.
By effectively turning off positional information---letting $a_{i} \rightarrow 0$ and setting $b_{i} = 0$---\cref{eqn:attention_pape} reduces to the standard attention score in \cref{eqn:attention_score}.
In this regime, PaPE no longer depends on position, enabling long-range interactions.

\textbf{PaPE decomposes attention.}
The decomposition in \cref{eqn:attention_pape} cleanly separates distinct contributions to the attention score:
$\langle a_{i}, \Delta r_{ij}^{\odot 2} \rangle$ encodes \textit{distance},
$\langle b_{i}, \Delta r_{ij} \rangle$ captures \textit{direction}, and 
$\langle q_{i}, k_{j} \rangle$ is the familiar \textit{semantic} term between tokens from \cref{eqn:attention_score}.
Although this separation is not one of our design principles, it is a valuable property for model analysis, see Appendix ~\ref{sec:model_analysis}.

\subsection[Compatibility with Efficient Attention Kernels (2)]{Compatibility with Efficient Attention Kernels \circlednum{2}}
\label{sec:efficient_pape}
The formulation in \cref{eqn:attention_pape} is not directly compatible with efficient attention kernels such as FlashAttention \cite{daoFlashAttentionFastMemoryEfficient2022,daoFlashAttention2FasterAttention2024}, because it explicitly depends on all pairwise relative positions.
To recover compatibility, we draw inspiration from RoPE \cite{suRoFormerEnhancedTransformer2024} and inject positional information directly into the dot product through separate position-aware transformations of the query and key.
These transformations must ensure that the resulting dot product exactly recovers \cref{eqn:attention_pape}.
That is, to guarantee that $\langle f_q(q_{i}, r_{i}), f_k(k_{j}, r_{j}) \rangle =$
\begin{equation}
    \label{eqn:efficient_pape}
    \langle a_{i}, \Delta r_{ij}^{\odot 2} \rangle + \langle b_{i}, \Delta r_{ij} \rangle + \langle q_{i}, k_{j} \rangle.
\end{equation}
Crucially, this preserves the standard attention computation.
Queries and keys are transformed independently using only absolute positions, and the attention kernel itself remains unchanged.
As a result, PaPE becomes plug-and-play compatible with efficient attention kernels.

Let $\oplus$ denote vector concatenation, $\otimes$ outer product, $M_i = W_p^T \mathrm{diag}(a_i) W_p$, and $\mathrm{flatten}(X)$ be row-major flattening of a matrix into a vector.
We can then transform the query as
\begin{equation}
    \label{eqn:f_q}
    f_q(q_{i}, r_{i}) =
    q_{i}
    \oplus r_i^T M_i r_i
    \oplus \mathrm{flatten}(M_i)
    \oplus - 2 M_i r_i
    \oplus W_p^T b_i
    \oplus - \langle W_p^T b_i, r_i \rangle ,
\end{equation}
and the key as
\begin{equation}
    \label{eqn:f_k}
    f_k(k_{j}, r_{j}) =
    k_{j}
    \oplus 1
    \oplus \mathrm{flatten}(r_j \otimes r_j)
    \oplus r_j
    \oplus r_j
    \oplus 1 .
\end{equation}

We prove in Appendix~\ref{sec:proof-efficient-pape} that these transformations exactly recover \cref{eqn:attention_pape}.

This construction preserves compatibility with efficient attention kernels with only a small overhead: the dimensionality of queries and keys increases by $p^2 + 2p + 2$.
This cost is moderate as $p$ is typically small at $p=2$ or $p=3$, increasing head sizes by 10 and 17, respectively.
We further measure and report efficiency in Appendix ~\ref{app:efficiency}.

\section{Experiments}
\label{sec:experiments}
We validate the main empirical claims by \circlednum{3} probing PaPE's ability to extrapolate in classification (\cref{sec:extrapolation}), and \circlednum{4} testing its generality on 8 datasets across 4 vision modalities (\cref{sec:generality}).
Beyond this, we strengthen the evaluation through parameter-matched experiments (\cref{sec:parameter_matched}), out-of-distribution generalization (\cref{sec:ood_generalization}), and an ablation study (\cref{sec:ablations}).

\subsection{Experimental Setup}
\label{sec:setup}
\begin{table*}[t]
    \centering
    \caption{
    \textbf{PaPE extrapolates well beyond the training resolution.}
    Training resolution is $224^2$ and evaluations are at resolutions up to $1024^2$.
    No additional training or modifications are applied.
    \textbf{Best} and \underline{next-best}.
    PaPE extrapolates better than the next-best encoding by up to $10.5$.
    }
    \resizebox{\linewidth}{!}{
    \begin{tabular}{l|ccccccccc}
        \toprule
        \textbf{Positional Encoding} & $224^2$ & $320^2$ & $384^2$ & $448^2$ & $512^2$ & $640^2$ & $768^2$ & $896^2$ & $1024^2$ \\
        \midrule
        nD-sincos \cite{wangTranslatingMathFormula2021} & 79.9 & 76.8 & 72.6 & 65.9 & 57.5 & 41.3 & 27.2 & 16.8 & 9.3 \\
        RoPE \cite{suRoFormerEnhancedTransformer2024} & 80.3 & 80.9 & 80.6 & 79.7 & 78.2 & 74.0 & 67.0 & 56.2 & 42.3 \\
        RoPE-Mixed \cite{heoRotaryPositionEmbedding2024} & 80.0 & 80.4 & 79.2 & 77.1 & 73.6 & 61.5 & 43.4 & 27.1 & 17.1 \\
        nD-ALiBi \cite{fullerCROMARemoteSensing2023} & 80.0 & 78.4 & 77.8 & 76.6 & 75.3 & 72.2 & 68.3 & 64.1 & 59.1 \\
        LookHere \cite{fullerLookHereVisionTransformers2024} & \underline{80.5} & \underline{81.4} & \underline{81.1} & \underline{80.4} & \underline{79.7} & \underline{77.5} & \underline{74.2} & \underline{69.6} & \underline{64.4} \\
        \midrule
        \textbf{PaPE (ours)} & \textbf{80.6} & \textbf{81.6} & \textbf{81.7} & \textbf{81.6} & \textbf{81.3} & \textbf{80.5} & \textbf{79.2} & \textbf{77.3} & \textbf{74.9} \\
        \textbf{PaPE-RI (ours)} & 80.0 & 80.2 & 79.2 & 78.1 & 76.9 & 73.6 & 69.7 & 64.6 & 58.6 \\
        \bottomrule
    \end{tabular}
    }
    \label{tab:extrapolation}
\end{table*}
\textbf{Baselines.}
We compare to five representative baselines:
nD-sincos \cite{wangTranslatingMathFormula2021} serves as a multi-dimensional generalization of the classic sinusoidal positional encoding of \citet{vaswaniAttentionAllYou2017}.
We use it to compare against an absolute positional encoding.
RoPE \cite{suRoFormerEnhancedTransformer2024} is a state-of-the-art and frequently used positional encoding, combining translation invariance with distance decay.
We specifically adopt the axial RoPE variant \cite{chuVisionLLaMAUnifiedLLaMA2024} to handle multiple position dimensions.
RoPE-Mixed \cite{heoRotaryPositionEmbedding2024} is included both because it is a common choice for adapting RoPE to vision and because it extends RoPE with directionality.
nD-ALiBi \cite{fullerCROMARemoteSensing2023} extends ALiBi \cite{pressTrainShortTest2021} to multiple dimensions using Euclidean distances, thereby achieving translation invariance, rotation invariance, and distance decay.
LookHere \cite{fullerLookHereVisionTransformers2024} imbues directionality to nD-ALiBi and loses rotation invariance in doing so.
It delivers high classification accuracy and strong extrapolation.
However, LookHere is only defined for images and does not admit a straightforward extension to other modalities, so we have to restrict it to image-based experiments.
We note that \citet{wangTranslatingMathFormula2021} in fact introduce a 2D variant of sinusoidal, and \citet{fullerCROMARemoteSensing2023} a 2D variant of ALiBi, but extending them to arbitrary dimensionality is straightforward.

\textbf{Datasets.}
We evaluate PaPE on 8 datasets that span diverse vision modalities and tasks, showcasing its broad applicability.
We consider three spatio-temporal datasets.
UCF101 \cite{soomroUCF101Dataset1012012} contains videos for action recognition of 101 actions.
DvsGesture \cite{amirLowPowerFully2017} and GEN1 \cite{detournemireLargeScaleEventbased2020} are event camera datasets.
Event streams are asynchronous, spatially sparse and temporally continuous, in contrast to synchronous, spatially dense, and temporally discrete videos.
DvsGesture targets action recognition of 11 actions and has high spatial sparsity.
GEN1 provides 39 hours of automotive driving for 2D object detection of cars and pedestrians and has medium spatial sparsity.
Thus, we get a rich variety of spatio-temporal data types.
ImageNet-1K \cite{deng2009imagenet,russakovskyImageNetLargeScale2015} offers 1M images for classification of 1K categories.
COCO \cite{linMicrosoftCOCOCommon2014} provides more than 200K images for 2D object detection of 80 categories.
Together, ImageNet-1K and COCO evaluate PaPE on large-scale image-based datasets.
ScanNet \cite{daiScanNetRichlyAnnotated3D2017} and ModelNet40 \cite{wu3DShapeNetsDeep2015} are 3D point cloud datasets.
ScanNet focuses on indoor scenes with widely distributed points and point-level semantic segmentation.
ModelNet40 contains densely sampled object shapes and targets object-level classification.
Finally, nuScenes \cite{caesarNuScenesMultimodalDataset2020} combines camera and LiDAR inputs for 3D object detection.
We use nuScenes to test PaPE in a challenging multimodal real-world setting.

\textbf{Models.}
The different modalities call for modality-specific Transformer variants.
We choose to use standard models that stay close to the original Transformer architecture.
For images, we adopt ViT \cite{dosovitskiyImageWorth16x162020} and equip it with a YOLOv10 \cite{wangYOLOv10RealTimeEndtoEnd2024a} head for object detection.  
For point clouds, we rely on Point Transformer V3 \cite{wuPointTransformerV32024}, optimized for 3D reasoning.  
For videos, we use ViViT \cite{arnabViViTVideoVision2021}, a dedicated video vision transformer.
For event cameras, we employ a ViT with spatio-temporal tokens obtained from Spiking Patches \cite{ohrstrom2025spiking} and re-use the YOLOv10 head for object detection on GEN1.
For nuScenes, we use UniTR \cite{wangUniTRUnifiedEfficient2023}, a unified multimodal transformer.

\textbf{Training.}
We implement all models in PyTorch \cite{Ansel_PyTorch_2_Faster_2024} using 16-bit mixed precision and optimize them with AdamW \cite{loshchilovDecoupledWeightDecay2019}.
All PaPE models are trained with $m = 50$.
The choice of $m = 50$ was made in preliminary experiments by tuning it on the validation split of CIFAR-10 \cite{krizhevsky2009learning}.
As such, $m$ has not been specifically tuned to any of the datasets that we consider in the paper. 
Models are validated at every epoch, and we select the checkpoint with the highest validation score for final evaluation on the test set.
All models are randomly initialized with uniform Kaiming initialization \cite{heDelvingDeepRectifiers2015}.
Crucially, for each dataset we fix the model size and all hyperparameters across positional encodings, so the only systematic difference between runs is the choice of positional encoding (up to training stochasticity).
This setup yields a fair comparison of the positional encodings.
See Appendix ~\ref{app:training_details} for details.

\subsection[How Well Does PaPE Extrapolate Beyond the Training Resolution? (3)]{How Well Does PaPE Extrapolate Beyond the Training Resolution? \circlednum{3}}
\label{sec:extrapolation}
We assess PaPE’s extrapolation ability (for classification) by evaluating how well it scales from its training resolution of $224^2$ up to $1024^2$ on ImageNet-1K.
Importantly, we do not change the models in any way for evaluation at higher resolutions.
\cref{tab:extrapolation} reports the raw results corresponding to the extrapolation plot shown in \cref{fig:fpf}b.

We find that PaPE extrapolates far better than any baseline, even outperforming LookHere, which is itself highly capable at extrapolating.
PaPE increases accuracy by up to 1.1 for resolutions up to $512^2$, and only falls below its training-resolution accuracy after $640^2$.
What is more, PaPE improves the accuracy by 10.5 over the next-best method (LookHere) at the highest resolution of $1024^2$.
Unlike the baselines, PaPE is context-aware.
This aligns with the findings of \citet{wangPositionalEncodingTokenAware2025} that adding a context-aware term significantly boosts RoPE’s extrapolation in language models. Taken together, these results suggest that context awareness has a role to play in extrapolation.

In contrast, PaPE-RI does not share the strong extrapolation.
This may be a consequence of the trade-off we made, sacrificing directionality to achieve rotation invariance.
Similarly, LookHere can be seen as an nD-ALiBi variant, enhanced with directionality.
In both cases, introducing directionality leads to better extrapolation, indicating that directionality is a key ingredient for extrapolation.

See additional details in Appendix ~\ref{app:extrapolation}, including an experiment (\cref{tab:extrapolation_best}) that confirms the superior extrapolation of PaPE, even when baselines are allowed to use position interpolation.

\subsection[Is PaPE a General Vision Positional Encoding? (4)]{Is PaPE a General Vision Positional Encoding? \circlednum{4}}
\label{sec:generality}
\begin{table*}[t]
    \centering
    \caption{\textbf{PaPE: a general vision positional encoding.}
    Comparison of PaPE and baselines across 8 diverse datasets. \textbf{Best} and \underline{next-best}.
    mAP is measured at 0.5:0.95:0.05 IoU intervals.
    PaPE matches the best baseline on 5 of 8 datasets and exceeds all on 2 datasets (UCF101 and GEN1), demonstrating the broad usefulness of PaPE across multiple vision modalities and tasks.
    }
    \resizebox{\linewidth}{!}{
    \begin{tabular}{l|ccc|cc|cc|c|c}
        \toprule
        \multicolumn{1}{c}{} & \multicolumn{3}{c}{\textbf{Spatio-temporal}} & \multicolumn{2}{c}{\textbf{Image}} & \multicolumn{2}{c}{\textbf{Point cloud}} & \multicolumn{1}{c}{\textbf{Multimodal}} & \multirow{3}{*}{\textbf{Average}} \\
        \cmidrule(lr){2-4} \cmidrule(lr){5-6} \cmidrule(lr){7-8} \cmidrule(lr){9-9}
        \multirow{2}{*}{\textbf{Positional Encoding}} & \textbf{UCF101} & \textbf{DvsGesture} & \textbf{GEN1} & \textbf{ImageNet-1K} & \textbf{COCO} & \textbf{ScanNet} & \textbf{ModelNet40} & \textbf{nuScenes} & \\
        & Acc. & Acc. & mAP & Acc. & mAP & mIoU & Acc. & mAP & \\
        \midrule
        nD-sincos \cite{wangTranslatingMathFormula2021} & 38.7 & \textbf{93.4} & 27.3 & 79.9 & 34.7 & \textbf{72.6} & 92.2 & 67.4 & 63.3 \\
        RoPE \cite{suRoFormerEnhancedTransformer2024} & \underline{43.9} & \underline{93.1} & \underline{33.6} & 80.3 & \underline{38.8} & \underline{71.7} & 93.0 & 68.3 & \underline{65.3} \\
        RoPE-Mixed \cite{heoRotaryPositionEmbedding2024} & 41.2 & 83.3 & 31.7 & 80.0 & 38.0 & 71.1 & 93.0 & 66.3 & 63.1 \\
        nD-ALiBi \cite{fullerCROMARemoteSensing2023} & 41.6 & 89.2 & 28.8 & 80.0 & 33.9 & 70.6 & \underline{93.2} & 68.5 & 63.2 \\
        LookHere \cite{fullerLookHereVisionTransformers2024} & - & - & - & \underline{80.5} & 37.3 & - & - & - & - \\
        \midrule
        \textbf{PaPE (ours)} & \textbf{49.5} & \textbf{93.4} & \textbf{34.8} & \textbf{80.6} & \textbf{38.9} & 71.0 & \textbf{93.3} & \underline{68.7} & \textbf{66.3} \\
        \textbf{PaPE-RI (ours)} & 42.2 & 90.6 & 28.5 & 80.0 & 34.3 & 69.3 & 92.7 & \textbf{68.9} & 63.3 \\
        \bottomrule
    \end{tabular}
    }
    \label{tab:generality}
\end{table*}
\cref{tab:generality} reports the results on the 8 datasets.
PaPE achieves the highest average score of 66.3---surpassing the next-best method, RoPE, by a margin of 1 point (65.3 vs. 66.3).
The score differences are small for some datasets; however, we still see that PaPE matches the best baseline on 5 datasets and clearly exceeds all baselines on 2 datasets, only underperforming on ScanNet.
These results validate that PaPE is a general vision positional encoding.
In contrast, PaPE-RI lags behind with a markedly lower average score of 63.3, highlighting the importance of directionality and the added flexibility in $a_i$ and $W_p$.
Breaking down the results reveals further insight:

\textbf{PaPE excels on spatio-temporal data.}
On UCF101, PaPE delivers the largest absolute accuracy gain among all datasets: 49.5 vs. 43.9 for RoPE.
We follow the official 3-fold cross-validation and report the mean accuracy (see Appendix ~\ref{app:ucf101_crossfold} for per-fold scores and significance test), highlighting the robustness of the result.
A potential reason for PaPE’s strong UCF101 performance is that many actions unfold very quickly, causing objects to move rapidly across frames---making directionality and context awareness matter greatly for understanding a fast moving object.
PaPE also leads on event camera datasets, outperforming RoPE by 1.2 mAP on GEN1 and tying for top accuracy with sinusoidal encodings on DvsGesture.
By contrast, RoPE-Mixed performs notably poorly in the spatio-temporal setting, especially on DvsGesture.
These results underscore how challenging spatio-temporal reasoning is—and how non-trivial it is for models to capture these relations effectively.

\textbf{PaPE is well-suited for images.}
On ImageNet-1K, PaPE has an accuracy of 80.6, matching that of LookHere at 80.5, and on COCO it likewise has a positive 0.1 difference in mAP compared to RoPE.
While the differences are small, these results show PaPE’s applicability to large-scale image datasets.

\textbf{PaPE is capable of handling multiple modalities.}
PaPE-RI delivers the strongest performance on nuScenes with 68.9 mAP, trailed by PaPE at 68.7, with nD-ALiBi close behind at 68.5.
This ranking is striking and points to an unexpectedly important role of rotation invariance in multimodal processing.
Mirroring our findings on spatio-temporal data, RoPE-Mixed again lags behind in this setting, showing how challenging it is to learn multimodal positional structure.

\textbf{When translation invariance matters.}
nD-sincos ranks as the lowest performing encoding on 4 of 8 datasets and trails as second-lowest on 2 of 8 datasets.
In particular, it consistently lags behind on object detection tasks (COCO, GEN1, and nuScenes).
While there are a few exceptions (ScanNet and DvsGesture), the pattern is clear:
the results suggest that translation invariance is often crucial, as nD-sincos is the only encoding in our study that lacks it.

\textbf{When absolute positional encoding matters.}
The ScanNet results reveal additional findings.
It is the only dataset where PaPE trails the baselines.
What is more, nD-sincos surpasses all others, including RoPE by 0.9 mIoU, suggesting that absolute positional information is important for ScanNet.
A potential explanation is that everyday household objects have characteristic sizes.
For example, a table typically sits about $1\,\mathrm{m}$ above the floor.
Purely relative encodings may fail to capture such relationships when the context in Point Transformer V3 does not span sufficiently distant points.
This explanation is supported by ModelNet40, where the translation invariant encodings surpass sinusoidal encodings by at least 0.8 accuracy.
Here, each sample is a single object with no surrounding scene, so absolute position carries less semantic weight.
In this setting, all methods achieve similar accuracy, yet PaPE remains the top performer, reinforcing it as a robust positional encoding.

\textbf{Rotation invariance matters less than hypothesized.}
Our point cloud experiments show that rotation invariant positional encodings do not improve over rotation variant alternatives.
This is evident from how neither rotation invariant encoding (nD-ALiBi and PaPE-RI) manages to outperform the rotation-variant encodings.
One possible explanation is that the rotation invariant positional encodings trade strict rotation invariance for less representation flexibility, and that decreased flexibility matters more than what they gain from invariance \cite{wilsonposition}.

\subsection{Does PaPE's Performance Stem from Additional Parameters?}
\label{sec:parameter_matched}
\begin{table*}
    \centering
    \caption{
    \textbf{Parameter-matched.}
    Increasing the parameters of the baselines to match the additional parameter counts introduced by PaPE does not change the ranking reported in \cref{tab:generality}.
    }
    \resizebox{\linewidth}{!}{
    \begin{tabular}{l|ccccc|cc}
        \toprule
        \textbf{Dataset} & \textbf{nD-sincos} & \textbf{RoPE} & \textbf{RoPE-Mixed} & \textbf{nD-ALiBi} & \textbf{LookHere} & \textbf{PaPE (ours)} & \textbf{PaPE-RI (ours)} \\
        \midrule
        UCF101 & 39.5 & \underline{45.2} & 41.3 & 41.9 & - & \textbf{49.5} & 41.8 \\
        DvsGesture & \textbf{93.4} & \underline{93.1} & 87.8 & 92.4 & - & \textbf{93.4} & \underline{93.1} \\
        ImageNet-1K & 79.6 & 80.2 & 80.1 & 80.0 & \underline{80.4} & \textbf{80.6} & 79.8 \\
        \bottomrule
    \end{tabular}
    }
    \label{tab:parameter-matched}
\end{table*}
PaPE increases the parameter count through $W_p$, $W_a$, and $W_b$, raising the question of whether its performance simply reflects greater model capacity.
We answer this using UCF101, DvsGesture, and ImageNet-1K by matching the baselines' parameter counts to PaPE's (see Appendix ~\ref{app:parameter-matched-details} for details).
Notably, this comparison favors the baselines, since whereas PaPE's extra parameters influence only positional terms, the matched baselines necessarily gain parameters that directly influence semantic representations as well.
Despite this, \cref{tab:parameter-matched} reports that matching parameters does not change the outcome observed in \cref{tab:generality}:
PaPE remains the highest scoring on UCF101 and ImageNet-1K, and tied with nD-sincos on DvsGesture, supporting PaPE as a strong positional encoding.
In addition, we report results on reducing non-PaPE parameters as well as extrapolation under parameter-matched settings in Appendix ~\ref{app:parameter-matched-details}.
PaPE maintains its advantage in all cases, further supporting that its high performance is due to improved positional encoding.

\subsection{Does PaPE Harm Out-of-distribution Generalization?}
\label{sec:ood_generalization}
\begin{table*}
    \centering
    \caption{
    \textbf{Out-of-distribution generalization.}
    PaPE does not degrade more than the baselines when evaluated on out-of-distribution versions of ImageNet.
    }
    \resizebox{\linewidth}{!}{
    \begin{tabular}{l|ccccc|cc}
        \toprule
        \textbf{Dataset} & \textbf{nD-sincos} & \textbf{RoPE} & \textbf{RoPE-Mixed} & \textbf{nD-ALiBi} & \textbf{LookHere} & \textbf{PaPE (ours)} & \textbf{PaPE-RI (ours)} \\
        \midrule
        ImageNetV2 \cite{pmlr-v97-recht19a} & 68.3 & \textbf{69.2} & 68.2 & 68.1 & \underline{68.9} & \textbf{69.2} & 68.8 \\
        ImageNet-Renditions \cite{Hendrycks_2021_ICCV} & 27.0 & \underline{29.2} & 28.2 & 27.2 & 28.6 & \textbf{29.4} & 26.8 \\
        ObjectNet \cite{barbuObjectNetLargescaleBiascontrolled2019} & 28.9 & \underline{31.0} & 28.5 & 29.2 & 29.8 & \textbf{31.1} & 29.4 \\
        \bottomrule
    \end{tabular}
    }
    \label{tab:ood_generalization}
\end{table*}
PaPE is context-aware through $W_a$ and $W_b$, making the positional encoding dependent on the feature space, and so raising the question if PaPE is harmful on out-of-distribution samples.
To assess this, we use the models trained on ImageNet-1K and evaluate them without additional training on ImageNetV2 \cite{pmlr-v97-recht19a}, ImageNet-Renditions \cite{Hendrycks_2021_ICCV}, and ObjectNet \cite{barbuObjectNetLargescaleBiascontrolled2019}.
ImageNetV2 has 10K images that are collected a decade after ImageNet.
ImageNet-Renditions has 30K image renditions (art, cartoons, sketches, etc.) of 200 classes from ImageNet-1K.
ObjectNet has 50K images of objects with varying rotations, backgrounds, and viewpoints.
ImageNetV2 seeks to stay close to the ImageNet domain whereas ImageNet-Renditions and ObjectNet focus on sampling from different input domains.
\cref{tab:ood_generalization} reports that PaPE is tied with RoPE as best performing on ImageNetV2 and it is the best performing on ImageNet-Renditions.
The differences in accuracy are small, but they confirm that PaPE does not degrade more than the baselines on out-of-distribution datasets.
We do not find PaPE to be harmful to out-of-distribution generalization.
We further confirm this for extrapolation in Appendix ~\ref{app:ood_extrapolation}.

\subsection{Ablation Study}
\label{sec:ablations}
\begin{table*}
    \centering
    \caption{
    \textbf{Ablations on ImageNet-1K.}
    All components of PaPE contribute to its accuracy as removing any component results in lower accuracy at the training resolution of $224^2$.
    $W_p$ does not influence the extrapolation performance, but we observe that all other components affect it, with distance decay (see $\langle a_i, \Delta r_{ij}^{\odot 2} \rangle$ and $a_i$ activation) particularly driving much of the high extrapolation.
    }
    \begin{tabular}{l|ccccccccc}
        \toprule
        \textbf{Ablation} & $224^2$ & $320^2$ & $384^2$ & $448^2$ & $512^2$ & $640^2$ & $768^2$ & $896^2$ & $1024^2$ \\
        \midrule
        PaPE & 79.2 & 80.3 & 80.1 & 79.5 & 78.9 & 77.5 & 75.4 & 72.9 & 69.8 \\
        - $\langle a_i, \Delta r_{ij}^{\odot 2} \rangle$ & 77.2 & 77.4 & 75.5 & 72.2 & 67.9 & 57.7 & 47.0 & 37.7 & 30.1 \\
        - $\langle b_i, \Delta r_{ij} \rangle$ & 78.7 & 79.3 & 78.5 & 77.4 & 76.1 & 73.1 & 69.5 & 64.9 & 59.5 \\
        - $W_a, W_b$ & 78.4 & 78.9 & 78.4 & 77.6 & 76.6 & 74.2 & 71.4 & 68.1 & 64.0 \\
        - $a_i$ activation & 78.6 & 79.7 & 79.4 & 78.4 & 76.9 & 71.9 & 63.5 & 50.7 & 36.0 \\
        - $W_p$ & 78.9 & 79.9 & 79.8 & 79.4 & 78.7 & 77.2 & 75.4 & 72.7 & 69.5 \\
        \bottomrule
    \end{tabular}
    \label{tab:ablations}
\end{table*}
\begin{table*}
\centering
\caption{
\textbf{Ablation efficiency.}
Measured on an RTX 4090 with a batch size of 1.
See Appendix ~\ref{app:efficiency} for detailed setup.
Context awareness ($W_a$, $W_b$) and $W_p$ are the main contributors of PaPE's moderate overhead.
Together with the results in \cref{tab:ablations}, this shows that we can remove $W_p$ to achieve high extrapolation without the moderate overhead.
}
\resizebox{\linewidth}{!}{
\begin{tabular}{lrrrrrrr}
\toprule
& & \multicolumn{3}{c}{\textbf{Training}} & \multicolumn{3}{c}{\textbf{Inference}} \\
\cmidrule(lr){3-5} \cmidrule(lr){6-8}
\textbf{Ablation} & \textbf{Params. (M)} & \textbf{VRAM (MB)} & \textbf{GFLOPS} & \textbf{Time (ms)} & \textbf{VRAM (MB)} & \textbf{GFLOPS} & \textbf{Time (ms)} \\
\midrule
PaPE & 97.5 & 1901 & 119.7 & 14.6 & 433 & 37.6 & 1.6 \\
- $\langle a_i, \Delta r_{ij}^{\odot 2} \rangle$ & 92.0 & 1791 & 112.8 & 13.7 & 388 & 35.3 & 1.5 \\
- $\langle b_i, \Delta r_{ij} \rangle$ & 92.0 & 1791 & 112.8 & 13.8 & 388 & 37.4 & 1.5 \\
- $W_a, W_b$ & 86.4 & 1673 & 106.6 & 12.9 & 363 & 35.3 & 1.4 \\
- $a_i$ activation & 97.5 & 1901 & 119.7 & 14.6 & 433 & 37.6 & 1.6 \\
- $W_p$ & 86.9 & 1679 & 106.7 & 13.1 & 365 & 35.4 & 1.5 \\
\bottomrule
\end{tabular}
}
\label{tab:ablation_efficiency}
\end{table*}
\begin{table*}
    \centering
    \caption{
    \textbf{Ablation of m on ImageNet-1K.}
    PaPE obtains the highest accuracy at $m=50$.
    The differences are small and a low $m$ may be preferable for increased efficiency.
    }
    \begin{tabular}{l|ccccccc}
        \toprule
        \textbf{m} & 2 & 4 & 8 & 16 & 32 & 50 & 64 \\
        \midrule
        \textbf{Accuracy} & 78.7 & 78.4 & 79.0 & 79.1 & 78.9 & 79.2 & 79.0 \\
        \bottomrule
    \end{tabular}
    \label{tab:ablation_m}
\end{table*}
We systematically ablate each component of PaPE to understand its contribution and to validate the instantiations of distance decay, directionality, and context awareness.
To ablate context awareness, we remove $W_a$ and $W_b$ and instead treat $a_i$ and $b_i$ as learned parameters, where the softplus activation is still applied to the $a$ coefficients.
They are learned distinctly for all layers and heads, but are shared between tokens such that $a_i = a_j$ and $b_i = b_j$ for any pair of $i$ and $j$.
This way, we maintain distance decay and directionality without context awareness.
To keep the study computationally feasible, we conduct all ablations on ImageNet-1K using a ViT-S instead of a ViT-B model size, and train for 150 epochs rather than 300.
\cref{tab:ablations} reports the ablation results.

\textbf{Ablation of principle instantiations.}
We see that PaPE has an accuracy of 79.2 and that all ablations result in lower accuracy.
Noticeably, removing any of the three instantiations of the design principles---distance decay ($\langle a_i, \Delta r_{ij}^{\odot 2} \rangle$), directionality ($\langle b_i, \Delta r_{ij} \rangle$), and context awareness ($W_a, W_b$)---causes an accuracy drop between $0.5$ and $2.0$ at the $224^2$ training resolution and from $5.8$ to $39.7$ at $1024^2$, validating the relevance of instantiating those principles in PaPE.
The ablations show that distance decay is a significant reason behind PaPE's extrapolation performance, as the accuracy drops by $39.7$ points without it.
Distance decay is further validated by how the model struggles to use the squared distance term when removing the softplus activation, resulting in accuracy decreases of $0.6$ at $224^2$ and $33.8$ at $1024^2$.
Directionality and context awareness also contribute to extrapolation as we observe respective drops in accuracy of $10.3$ and $5.8$ at $1024^2$ without them.

\textbf{Ablation of $W_p$ and $m$.}
Although its impact is modest at a $0.3$ difference in accuracy, adding $W_p$ contributes positively to PaPE.
$W_p$ does not influence extrapolation since the accuracy drops at all resolutions range from $0$ to $0.4$ without $W_p$, aligned with the training drop of $0.3$.
We also report ablation efficiency in \cref{tab:ablation_efficiency} and find that $W_p$ is a significant contributor to PaPE's moderate overhead.
This matters because we can remove $W_p$ to reduce the overhead significantly while preserving the extrapolation performance.
\cref{tab:ablation_m} further confirms the choice of $m=50$ as it obtains the highest accuracy of 79.2.
We note though that the difference is small and that decreasing $m$ improves the efficiency.
In practice, $m=8$ provides a good trade-off between accuracy and efficiency, evident from the efficiency reported for various $m$ in Appendix ~\ref{app:efficiency}.
Additionally, Appendix ~\ref{app:m_extrapolation} shows that $m$ has low influence on extrapolation, meaning the choice of $m$ is of little importance in this regard.

\section{Limitations \& Future Work}
\label{sec:limitations}
PaPE’s main limitation is its moderate overhead (Appendix ~\ref{app:efficiency}), dependent on $m$.
Future work will focus on how to lower $m$ in practice, or on removing $W_p$ altogether and thereby keeping $m = p$, which is small for all modalities and tasks considered here.
Additionally, we want to integrate PaPE directly into efficient attention kernels to further decrease the overhead.
Finally, an interesting direction is to explore parameter-efficient fine-tuning (PEFT) using the PaPE parameters for position-based PEFT.

\section{Conclusions}
\label{sec:conclusion}
We introduce Parabolic Positional Encoding (PaPE), a principled, vision-centric positional encoding.
PaPE, along with its rotation invariant version PaPE-RI, is derived from principles identified in prior research and is compatible with efficient attention kernels.
The experiments demonstrate that PaPE exhibits exceptional classification extrapolation capabilities, substantially outperforming all baselines by at least 10.5 at resolution $1024^2$.
We evaluate PaPE on 8 datasets spanning videos, event cameras, images, point clouds, and multimodal settings.
The results validate PaPE as a general vision positional encoding: it raises the average score by 1 over RoPE while matching the best baseline on 5 datasets and exceeding all baselines on 2 datasets.
Thus, PaPE paves a new path for vision positional encoding, and we anticipate that its underlying principles will motivate future research.

\begin{ack}
This work has been supported by Innovation Fund Denmark through the project ``Safety and Autonomy for Vehicles in Agriculture (SAVA)", 2105-00013A.
\end{ack}

\bibliography{main}
\bibliographystyle{abbrvnat}

\appendix

\section{Proofs}
\label{sec:proofs}

\subsection{PaPE-RI is Rotation Invariant}
\label{sec:proof-papeRI}

By the definition of PaPE-RI, assume $b_{i} = 0$, $a_{i\ell} = \alpha_i \in \mathbb{R}_{<0}$, and $W_p = w_p I_p$ with $w_p \in \mathbb{R}$, where $I_p$ represents the identity matrix of shape $p\times p$.

Let $\Delta r_{ij} = W_p  \cdot (r_{j} - r_{i})$ and $\Delta r_{ij}^\prime = W_p \cdot (Rr_{j} - Rr_{i})$ for $R \in \textrm{SO}(p)$.

In this proof, we show that restricting the model to the conditions assumed above guarantees that $A_{ij} {=} A_{ij}^\prime$, where entries $A_{ij}$ depend on $\Delta r_{ij}$, and entries $A_{ij}^\prime$ depend respectively on $\Delta r_{ij}^\prime$.
This satisfies Definition ~\ref{def:rotation_invariance}.

Inserting the conditions in the position-dependent terms leads to the simplified expression,
\begin{align}  
    & \left( \sum_{\ell=1}^{m} a_{i\ell}  \Delta r_{ij\ell}^{2} + b_{i\ell}  \Delta r_{ij\ell} \right) \\
    & = \sum_{\ell=1}^{m} a_{i\ell} \Delta r_{ij\ell}^{2} \quad \text{using} \quad b_{i\ell} = 0\\
    & = \alpha_i \sum_{\ell=1}^{m} \Delta r_{ij\ell}^{2} \quad \text{using} \quad a_{i\ell} = \alpha_i\\
    & = \alpha_i \Delta r_{ij}^T \Delta r_{ij}.
\end{align}

The equivalence of the position-dependent terms ($\Delta r_{ij}^{\prime T} \Delta r_{ij}^\prime = \Delta r_{ij}^T \Delta r_{ij}$) proves the rotation invariance of PaPE-RI:
\begin{align}
    & \Delta r_{ij}^{\prime T} \Delta r_{ij}^\prime \\
    & = (W_p \cdot R(r_{j} - r_{i}))^T(W_p \cdot R(r_{j} - r_{i})) \\
    & = w_p^2 (r_j - r_i)^TR^TR(r_j - r_i) \quad \text{using} \quad W_p = w_p I_p \\
    & = w_p^2 (r_j - r_i)^T(r_j - r_i) \quad \text{using} \quad R^TR = I_p, R \in \textrm{SO}(p) \\
    & = \Delta r_{ij}^T \Delta r_{ij} \\
    & \implies A_{ij} = A_{ij}^\prime \quad \square.
\end{align}

\subsection{PaPE is Compatible with Efficient Attention Kernels}
\label{sec:proof-efficient-pape}
We claim in \cref{sec:efficient_pape} that
\begin{equation}
    \label{eqn:efficient_pape_claim}
    \langle f_q(q_{i}, r_{i}), f_k(k_{j}, r_{j}) \rangle = \langle a_{i}, \Delta r_{ij}^{\odot 2} \rangle + \langle b_{i}, \Delta r_{ij} \rangle + \langle q_{i}, k_{j} \rangle.
\end{equation}
Recall that $M_i=W_p^T \mathrm{diag}(a_i) W_p$ and that we define in Equations~\eqref{eqn:f_q} and \eqref{eqn:f_k} $f_q$ and $f_k$ as

\bgroup
\setlength\tabcolsep{3.5pt}
\begin{tabular}{cccccccccccccc}
     $f_q(q_{i}, r_{i})$ & $=$ & $q_{i}$ & $\oplus$ & $r_i^T M_i r_i$ & $\oplus$ & $\mathrm{flatten}(M_i)$ & $\oplus$ & $- 2 M_i r_i$ & $\oplus$ & $W_p^T b_i$ & $\oplus$ & $- \langle W_p^T b_i, r_i \rangle$ \\
     $f_k(k_{j}, r_{j})$ & $=$ & $k_{j}$ & $\oplus$ & 1 & $\oplus$ & $\mathrm{flatten}(r_j \otimes r_j)$ & $\oplus$ & $r_j$ & $\oplus$ & $r_j$ & $\oplus$ & 1
\end{tabular}
\egroup

Completing the dot product block by block shows that $\langle f_q(q_{i}, r_{i}), f_k(k_{j}, r_{j}) \rangle = $
\begin{equation}
    \langle q_i, k_j \rangle + r_i^T M_i r_i + \langle \mathrm{flatten}(M_i), \mathrm{flatten}(r_j \otimes r_j) \rangle - \langle 2 M_i r_i, r_j \rangle + \langle W_p^T b_i, r_j \rangle - \langle W_p^T b_i, r_i \rangle.
\end{equation}

As such, we prove the claim by proving the following equations
\begin{align}
    \label{eqn:proof_quadratic}
    \langle a_{i}, \Delta r_{ij}^{\odot 2} \rangle &= r_i^T M_i r_i + \langle \mathrm{flatten}(M_i), \mathrm{flatten}(r_j \otimes r_j) \rangle - \langle 2 M_i r_i, r_j \rangle \\
    \label{eqn:proof_linear}
    \langle b_{i}, \Delta r_{ij} \rangle &= \langle W_p^T b_i, r_j \rangle - \langle W_p^T b_i, r_i \rangle
\end{align}
from which \cref{eqn:efficient_pape_claim} directly follows.

\textbf{Proof of quadratic terms (\cref{eqn:proof_quadratic}).}

First, we see that $M$ is symmetric because
\begin{equation}
    \left[ W_p^T \mathrm{diag}(a_i) W_p \right]_{jk} = \sum_{\ell=1}^{m} a_\ell w_{\ell j} w_{\ell k},
\end{equation}
and $w_{\ell j} \cdot w_{\ell k} = w_{\ell k} \cdot w_{\ell j}$.
Therefore, $M_{jk} = M_{kj}$, making $M$ symmetric.

It also holds that
\begin{equation}
    2 r_j^T M_i r_i = r_i^T M_i r_j + r_j^T M_i r_i
\end{equation}
because $M_i$ is symmetric.

We also note that
\begin{equation}
    \langle \mathrm{flatten}(M_i), \mathrm{flatten}(r_j \otimes r_j) \rangle = r_j^T M_i r_j.
\end{equation}

Together, we can write the right hand side of \cref{eqn:proof_quadratic} as
\begin{equation}
    r_i^T M_i r_i + r_j^T M_i r_j - r_i^T M_i r_j - r_j^T M_i r_i,
\end{equation}
which simplifies to
\begin{align}
    r_i^T M_i (r_i - r_j) + r_j^T M_i (r_j - r_i) &= - r_i^T M_i (r_j - r_i) + r_j^T M_i (r_j - r_i) \\
    &= (r_j - r_i)^T M_i (r_j - r_i)
\end{align}
Substituting with $M_i = W_p^T \mathrm{diag}(a_i) W_p$ completes the proof:
\begin{align}
    (r_j - r_i)^T M_i (r_j - r_i)
    &= (r_j - r_i)^T W_p^T \mathrm{diag}(a_i) W_p (r_j - r_i) \\
    &= \left( (r_j - r_i)^T W_p^T \right) \mathrm{diag}(a_i) \left( W_p (r_j - r_i) \right) \\
    &= \Delta r_{ij}^T \mathrm{diag}(a_i) \Delta r_{ij} \\
    &= \langle a_{i}, \Delta r_{ij}^{\odot 2} \rangle \quad \square.
\end{align}

\textbf{Proof of linear terms (\cref{eqn:proof_linear}).}

We verify \cref{eqn:proof_linear} below:
\begin{align}
    \langle W_p^T b_i, r_j \rangle - \langle W_p^T b_i, r_i \rangle
    &= \langle W_p^T b_i, r_j - r_i \rangle \\
    &= \left( W_p^T b_i \right)^T (r_j - r_i) \\
    &= \left( b_i^T W_p \right) (r_j - r_i) \\
    &= b_i^T \left( W_p (r_j - r_i) \right) \\
    &= b_i^T \Delta r_{ij} \\
    &= \langle b_i, \Delta r_{ij} \rangle \quad \square.
\end{align}

\section{Generalizations of PaPE}
We motivate and discuss generalizations of PaPE.

\subsection{Polynomial Generalizations of PaPE} \label{sec:generalization}
PaPE can be viewed as the first instance of a larger family of multivariate polynomials on $\Delta r_{ij}$ with context-dependent coefficients.
Although we leave this direction unexplored in the current work, we are eager to investigate other members of this function family in future work.
We elaborate on this perspective here and describe polynomial generalizations of PaPE.

Let $x_i \in \mathbb{R}^d$ and $x_j \in \mathbb{R}^d$ be two tokens at positions $r_i \in \mathbb{R}^p$ and $r_j \in \mathbb{R}^p$.  
An expressive family of continuous, context-aware, and translation invariant functions are polynomials of the difference $(r_j-r_i) \in \mathbb{R}^p$,
\begin{align}
    g(x_i, r_i, x_j, r_j) = \sum_{n=0}^{N} a_n(x_i, x_j)(r_j - r_i)^{\odot n},
\end{align}
with coefficients $a_n \in \mathbb{R}^{1\times p}$ and where $(\cdot)^{\odot n}$ represents the element-wise exponentiation.

For compatibility with efficient attention kernels, the function $g$ needs to be representable as a dot product of finite feature vectors that each only depend on the position and content of either of the two tokens.
Since $(r_j-r_i)^n$ can be expanded to sums of monomials $r_i^lr_j^{n-l}$, the expression is separable into feature vectors, as long as the coefficients $a_n$ are of the type $a_n = h_q(x_i) h_k(x_j)$ for functions $h_q, h_k$.
This results in terms of type $\langle h_q(x_i) r_i^l , r_j^{n-l} h_k(x_j)\rangle$.
Note that in PaPE we set $h_k(x_j)=1$, leading to coefficients of type $a_n(x_i, x_j) = h_q(x_i)$.

A further generalization of PaPE is into the multivariate polynomials. 
To simplify notation, let $x = (r_j - r_i) = (x_1,x_2,\dots,x_p)$. 
Multivariate polynomials are finite sums of terms of the type $c_\alpha x_1^{\alpha_1} x_2^{\alpha_2} \dots x_p^{\alpha_p}$, where $c_\alpha \in \mathbb{R}$, and $\alpha = (\alpha_1,\alpha_2, \dots, \alpha_p) \in \mathbb{N}^p$ is a multi-index.
The degree $N$ of the multivariate polynomial is the largest sum of exponents corresponding to non-zero coefficients, i.e., $N = \max\{ \sum_{i=1}^{p}\alpha_i \mid c_\alpha \neq 0 \}$.

This broader family can represent terms that cannot be represented in the former, e.g., bilinear forms $(r_j -r_i)^TB(r_j -r_i)$ for any $B\in \mathbb{R}^{p\times p}$.
Similar to the former case, multivariate polynomials can also be separated into feature vectors for efficient attention kernel compatibility, as long as coefficients $c_\alpha$ are of the type $h_q(q_i)h_k(k_j)$.

\subsection{Relaxations of Translation and Rotation Invariance}
\label{app:relaxations}
We have presented PaPE and PaPE-RI.
PaPE is translation invariant and PaPE-RI is also rotation invariant.
We discussed in \cref{sec:generality} how absolute positions can be useful in some edge cases, and we explicitly stated rotation invariance as a task-specific principle (\cref{sec:design_principles}).
Therefore, we here propose extensions of PaPE to relax translation and rotation invariance when doing so may be useful.
Note that these extensions are untested, but may inspire future work.

\textbf{Relaxing translation invariance.}
We include a position-dependent term that turns \cref{eqn:attention_pape} into
\begin{equation}
    \label{eqn:trans_inv_relax}
    S_{ij} = \langle a_{i}, \Delta r_{ij}^{\odot 2} \rangle + \langle b_{i}, \Delta r_{ij} \rangle + \langle q_{i}, k_{j} \rangle + \langle c_i, W_p r_i \rangle ,
\end{equation}
where $c_i = W_c x_i$ with $W_c \in \mathbb{R}^{m \times d}$ being learnable.
The final term breaks translation invariance.
We can further relax its influence through a loss that encourages the norm of $c_i$ to be close to 0, making it approximately translation invariant, but enabling the model to still use absolute positions.

\textbf{Relaxing rotation invariance.}
We can relax PaPE to be approximately rotation invariant by regularizing $W_p$ towards the constrained form used in PaPE-RI and regularizing $b_i$ towards 0, recovering PaPE-RI in the limit of strong regularization and unconstrained PaPE without regularization.
Such regularization could, for example, be achieved with an auxiliary loss.

\section{Additional Experiments \& Results}
This section reports experiments and results that are informative to understand PaPE's performance, but does not fit within the main text.

\subsection{Baselines}
\label{app:more_baselines}
\begin{table*}
    \centering
    \caption{\textbf{Additional baselines.}
    PaPE confirms its high generality when compared to context-aware baselines and learned positional embeddings from ViT.
    }
    \begin{tabular}{l|rrr|r}
        \toprule
        \textbf{Dataset} & \textbf{CPVT} & \textbf{CvT} & \textbf{ViT-PE} & \textbf{PaPE (ours)} \\
        \midrule
        UCF101 & - & - & 38.3 $\pm$ 3.1 & \textbf{49.5} $\pm$ 1.6 \\
        ImageNet-1K & 79.2 & 78.9 & 79.2 & \textbf{80.6} \\
        COCO & 32.1 & 34.1 & 14.7 & \textbf{38.9} \\
        \bottomrule
    \end{tabular}
    \label{tab:more_baselines_results}
\end{table*}
\begin{table*}
    \centering
    \caption{
    \textbf{Extrapolation with additional baselines.}
    PaPE remains best at extrapolation by large margins ranging from $17.6$ to $21.0$ at resolution $1024^2$.
    }
    \begin{tabular}{l|rrrrrrrrr}
        \toprule
        \textbf{Positional Encoding} & $224^2$ & $320^2$ & $384^2$ & $448^2$ & $512^2$ & $640^2$ & $768^2$ & $896^2$ & $1024^2$ \\
        \midrule
        CPVT & 79.2 & 79.6 & 79.1 & 78.0 & 76.6 & 73.4 & 68.5 & 62.9 & 56.6 \\
        CvT & 78.9 & 79.5 & 79.0 & 78.2 & 77.1 & 73.7 & 68.4 & 61.4 & 53.9 \\
        ViT-PE & 79.2 & 79.1 & 78.1 & 76.6 & 75.1 & 71.7 & 67.4 & 62.6 & 57.3 \\
        \midrule
        \textbf{PaPE (ours)} & \textbf{80.6} & \textbf{81.6} & \textbf{81.7} & \textbf{81.6} & \textbf{81.3} & \textbf{80.5} & \textbf{79.2} & \textbf{77.3} & \textbf{74.9} \\
        \bottomrule
    \end{tabular}
    \label{tab:more_baselines_extrapolation}
\end{table*}
Here we compare against additional baselines.
First, we include the learned positional embeddings from ViT \cite{dosovitskiyImageWorth16x162020} and denote it ViT-PE.
This gives a comparison to another absolute positional encoding besides nD-sincos.
We also include context-aware baselines: CPVT \cite{chuConditionalPositionalEncodings2022} and CvT \cite{wuCvTIntroducingConvolutions2021}.
ViT-PE is only compatible with discrete grids, meaning we can only evaluate it on UCF101, ImageNet-1K, and COCO.
We follow the usual approach of using bicubic interpolation on the 2D grid when extrapolating with ViT-PE.
CPVT and CvT further impose 2D constraints due to using 2D convolutions, restricting their evaluation to ImageNet-1K and COCO.

\cref{tab:more_baselines_results} reports the findings of the additional baselines.
We find that PaPE consistently obtains higher accuracy than the additional baselines.
This validates PaPE's high generality even further.
\cref{tab:more_baselines_extrapolation} also validates this for extrapolation.
We find PaPE to remain a strong positional encoding when compared to ViT-PE, CPVT, and CvT.

\subsection{Extrapolation with and without Position Interpolation}
\label{app:extrapolation}
Here, we further elaborate on the extrapolation performance of PaPE and the baseline methods.

\textbf{Without position interpolation.}
The main extrapolation experiment evaluates how the methods perform without position interpolation.
Concretely, during training we only see positions in the range $[1, 14]$ because we train at a resolution of $224^2$ with a patch size of 16, since $224 / 16 = 14$.
At higher resolutions, however, this expands up to $[1, 64]$, since $1024 / 16 = 64$.
As previously mentioned, we see in \cref{tab:extrapolation} that PaPE excels in this setting.

\textbf{With position interpolation.}
Prior works find that sinusoidal and RoPE-based positional encodings extrapolate better with position interpolation \cite{chenExtendingContextWindow2023,dingLongRoPEExtendingLLM2024}.
It works by mapping the test-time position range back into the original training range.
We do this by scaling the positions by a factor of $224 / R$, where $R$ is the target resolution.
For example, at $R = 448$, each position is multiplied by $0.5$, compressing the effective positions back into the $[1, 14]$ training range.
\begin{table*}
    \centering
    \caption{
    \textbf{Extrapolation results with position interpolation.}
    The brackets indicate the increase (\textcolor{increase}{green}) or decrease (\textcolor{red}{red}) compared to the corresponding configuration without interpolation in \cref{tab:extrapolation}. 
    \textbf{Best} and \underline{next-best}.
    nD-sincos benefits greatly from interpolation whereas RoPE and RoPE-Mixed benefit from resolution $640^2$ and above.
    Position interpolation is detrimental to all of the attention bias methods (nD-ALiBi, LookHere, PaPE, and PaPE-RI).
    }
    \resizebox{\linewidth}{!}{
    \begin{tabular}{l|lllllllll}
        \toprule
        \textbf{Positional Encoding} & $224^2$ & $320^2$ & $384^2$ & $448^2$ & $512^2$ & $640^2$ & $768^2$ & $896^2$ & $1024^2$ \\
        \midrule
nD-sincos \cite{wangTranslatingMathFormula2021} & 79.9 & 79.9 \small{\textcolor{increase}{(+3.1)}} & 78.8 \small{\textcolor{increase}{(+6.2)}} & 77.4 \small{\textcolor{increase}{(+11.5)}} & 75.8 \small{\textcolor{increase}{(+18.3)}} & 71.8 \small{\textcolor{increase}{(+30.5)}} & 66.9 \small{\textcolor{increase}{(+39.7)}} & 61.3 \small{\textcolor{increase}{(+44.5)}} & 56.0 \small{\textcolor{increase}{(+46.7)}} \\
RoPE \cite{suRoFormerEnhancedTransformer2024} & 80.3 & 80.6 \small{\textcolor{red}{(-0.3)}} & \underline{80.0} \small{\textcolor{red}{(-0.6)}} & \textbf{78.9} \small{\textcolor{red}{(-0.8)}} & \textbf{77.6} \small{\textcolor{red}{(-0.6)}} & \textbf{74.5} \small{\textcolor{increase}{(+0.5)}} & \textbf{69.9} \small{\textcolor{increase}{(+2.9)}} & \textbf{64.8} \small{\textcolor{increase}{(+8.6)}} & \textbf{59.8} \small{\textcolor{increase}{(+17.5)}} \\
RoPE-Mixed \cite{heoRotaryPositionEmbedding2024} & 80.0 & 72.9 \small{\textcolor{red}{(-7.5)}} & 72.1 \small{\textcolor{red}{(-7.1)}} & 73.0 \small{\textcolor{red}{(-4.1)}} & 67.8 \small{\textcolor{red}{(-5.8)}} & 63.5 \small{\textcolor{increase}{(+2.0)}} & 56.1 \small{\textcolor{increase}{(+12.7)}} & 49.5 \small{\textcolor{increase}{(+22.4)}} & 42.9 \small{\textcolor{increase}{(+25.8)}} \\
nD-ALiBi \cite{fullerCROMARemoteSensing2023} & 80.0 & 77.9 \small{\textcolor{red}{(-0.5)}} & 77.2 \small{\textcolor{red}{(-0.6)}} & 76.0 \small{\textcolor{red}{(-0.6)}} & 74.4 \small{\textcolor{red}{(-0.9)}} & 70.6 \small{\textcolor{red}{(-1.6)}} & 65.4 \small{\textcolor{red}{(-2.9)}} & 59.8 \small{\textcolor{red}{(-4.3)}} & 53.9 \small{\textcolor{red}{(-5.2)}} \\
LookHere \cite{fullerLookHereVisionTransformers2024} & \underline{80.5} & \textbf{80.9} \small{\textcolor{red}{(-0.5)}} & \textbf{80.2} \small{\textcolor{red}{(-0.9)}} & \textbf{78.9} \small{\textcolor{red}{(-1.5)}} & \underline{77.4} \small{\textcolor{red}{(-2.3)}} & \underline{73.6} \small{\textcolor{red}{(-3.9)}} & \underline{68.8} \small{\textcolor{red}{(-5.4)}} & \underline{63.7} \small{\textcolor{red}{(-5.9)}} & \underline{59.0} \small{\textcolor{red}{(-5.4)}} \\
\midrule
\textbf{PaPE (ours)} & \textbf{80.6} & \underline{80.7} \small{\textcolor{red}{(-0.9)}} & 79.6 \small{\textcolor{red}{(-2.1)}} & \underline{78.1} \small{\textcolor{red}{(-3.5)}} & 76.5 \small{\textcolor{red}{(-4.8)}} & 72.0 \small{\textcolor{red}{(-8.5)}} & 66.4 \small{\textcolor{red}{(-12.8)}} & 60.4 \small{\textcolor{red}{(-16.9)}} & 54.8 \small{\textcolor{red}{(-20.1)}} \\
\textbf{PaPE-RI (ours)} & 80.0 & 80.3 \small{\textcolor{increase}{(+0.1)}} & 79.4 \small{\textcolor{increase}{(+0.2)}} & \underline{78.1} \small{\textcolor{black}{(0.0)}} & 76.4 \small{\textcolor{red}{(-0.5)}} & 72.2 \small{\textcolor{red}{(-1.4)}} & 66.5 \small{\textcolor{red}{(-3.2)}} & 59.7 \small{\textcolor{red}{(-4.9)}} & 52.8 \small{\textcolor{red}{(-5.8)}} \\
        \bottomrule
    \end{tabular}
    }
    \label{tab:extrapolation_interpolated}
\end{table*}
\cref{tab:extrapolation_interpolated} shows how PaPE and the baselines perform with position interpolation.
We clearly see how nD-sincos in particular is greatly improved with position interpolation.
RoPE and RoPE-Mixed drop for resolutions below $640^2$, but greatly increase their accuracy from there and up.
These results confirm that position interpolation is useful for sinusoidal encodings and partially useful for RoPE-based encodings.
We also find that all four attention bias positional encodings, including PaPE, are negatively affected by position interpolation.

\textbf{Best-performing configuration.}
Although position interpolation partially mitigates the shortcomings of sinusoidal and RoPE-based encodings, there is still no encoding that approaches the extrapolation performance of PaPE \emph{without} position interpolation.
\begin{table*}
    \centering
    \caption{
    \textbf{Extrapolation results for best-performing configuration per positional encoding.}
    With position interpolation: nD-sincos, RoPE, and RoPE-Mixed.
    Without: nD-ALiBi, LookHere, PaPE, and PaPE-RI.
    \textbf{Best} and \underline{next-best}.
    Although position interpolation is detrimental to PaPE, it remains the best encoding for classification extrapolation---even when applying position interpolation to the encodings that benefit from it.
    }
    \resizebox{\linewidth}{!}{
    \begin{tabular}{l|ccccccccc}
        \toprule
        \textbf{Positional Encoding} & $224^2$ & $320^2$ & $384^2$ & $448^2$ & $512^2$ & $640^2$ & $768^2$ & $896^2$ & $1024^2$ \\
        \midrule
nD-sincos \cite{wangTranslatingMathFormula2021} & 79.9 & 79.9 & 78.8 & 77.4 & 75.8 & 71.8 & 66.9 & 61.3 & 56.0 \\
RoPE \cite{suRoFormerEnhancedTransformer2024} & 80.3 & 80.6 & 80.0 & 78.9 & 77.6 & 74.5 & 69.9 & 64.8 & 59.8 \\
RoPE-Mixed \cite{heoRotaryPositionEmbedding2024} & 80.0 & 72.9 & 72.1 & 73.0 & 67.8 & 63.5 & 56.1 & 49.5 & 42.9 \\
nD-ALiBi \cite{fullerCROMARemoteSensing2023} & 80.0 & 78.4 & 77.8 & 76.6 & 75.3 & 72.2 & 68.3 & 64.1 & 59.1 \\
LookHere \cite{fullerLookHereVisionTransformers2024} & \underline{80.5} & \underline{81.4} & \underline{81.1} & \underline{80.4} & \underline{79.7} & \underline{77.5} & \underline{74.2} & \underline{69.6} & \underline{64.4} \\
\midrule
\textbf{PaPE (ours)} & \textbf{80.6} & \textbf{81.6} & \textbf{81.7} & \textbf{81.6} & \textbf{81.3} & \textbf{80.5} & \textbf{79.2} & \textbf{77.3} & \textbf{74.9} \\
\textbf{PaPE-RI (ours)} & 80.0 & 80.2 & 79.2 & 78.1 & 76.9 & 73.6 & 69.7 & 64.6 & 58.6 \\
        \bottomrule
    \end{tabular}
    }
    \label{tab:extrapolation_best}
\end{table*}
This is evident in \cref{tab:extrapolation_best}, which reports extrapolation scores for the best-performing configuration of each positional encoding.
In this comparison, nD-sincos, RoPE, and RoPE-Mixed rely on interpolation, whereas nD-ALiBi, LookHere, PaPE, and PaPE-RI do not.
We observe in particular that PaPE remains the strongest encoding across all resolutions, with LookHere consistently ranking as the next-best method for extrapolation.
As a result, the 10.5 improvement at resolution $1024^2$ persists even after taking position interpolation into account.

\subsection{Out-of-distribution Extrapolation}
\label{app:ood_extrapolation}
\begin{table*}
    \centering
    \caption{
    \textbf{Extrapolation on ImageNetV2.}
    PaPE maintains its rank as the \textbf{best} positional encoding at extrapolating, increasing the accuracy at $1024^2$ by 9.9 over the \underline{next-best} (LookHere).
    }
    \resizebox{\linewidth}{!}{
    \begin{tabular}{l|rrrrrrrrr}
        \toprule
        \textbf{Positional Encoding} & $224^2$ & $320^2$ & $384^2$ & $448^2$ & $512^2$ & $640^2$ & $768^2$ & $896^2$ & $1024^2$ \\
        \midrule
nD-sincos \cite{wangTranslatingMathFormula2021} & 68.3 & 64.2 & 58.1 & 50.7 & 42.3 & 27.1 & 16.6 & 10.1 & 5.6 \\
RoPE \cite{suRoFormerEnhancedTransformer2024} & \textbf{69.2} & 70.3 & 69.7 & 67.9 & 65.9 & 60.1 & 51.8 & 41.2 & 29.1 \\
RoPE-Mixed \cite{heoRotaryPositionEmbedding2024} & 68.2 & 69.2 & 67.7 & 64.3 & 59.5 & 44.2 & 28.3 & 17.0 & 10.4 \\
nD-ALiBi \cite{fullerCROMARemoteSensing2023} & 68.1 & 69.4 & 67.7 & 65.7 & 63.9 & 59.4 & 54.6 & 49.1 & 43.4 \\
LookHere \cite{fullerLookHereVisionTransformers2024} & \underline{68.9} & \underline{70.8} & \underline{70.2} & \underline{69.5} & \underline{68.3} & \underline{65.5} & \underline{61.0} & \underline{56.5} & \underline{51.6} \\
\midrule
\textbf{PaPE (ours)} & \textbf{69.2} & \textbf{70.9} & \textbf{71.2} & \textbf{70.6} & \textbf{70.1} & \textbf{68.9} & \textbf{66.7} & \textbf{64.5} & \textbf{61.5} \\
\textbf{PaPE-RI (ours)} & 68.8 & 69.3 & 67.9 & 66.2 & 64.3 & 59.5 & 54.3 & 48.8 & 43.1 \\
        \bottomrule
    \end{tabular}
    }
    \label{tab:extrapolation_imagenetv2}
\end{table*}
\begin{table*}
    \centering
    \caption{
    \textbf{Extrapolation on ImageNet-Renditions.}
    PaPE maintains its rank as the \textbf{best} positional encoding at extrapolating, increasing the accuracy at $1024^2$ by 8.2 over the \underline{next-best} (LookHere).
    }
    \resizebox{\linewidth}{!}{
    \begin{tabular}{l|rrrrrrrrr}
        \toprule
        \textbf{Positional Encoding} & $224^2$ & $320^2$ & $384^2$ & $448^2$ & $512^2$ & $640^2$ & $768^2$ & $896^2$ & $1024^2$ \\
        \midrule
nD-sincos \cite{wangTranslatingMathFormula2021} & 27.0 & 18.4 & 13.3 & 8.7 & 5.4 & 2.5 & 1.3 & 0.9 & 0.4 \\
RoPE \cite{suRoFormerEnhancedTransformer2024} & \underline{29.2} & \underline{26.8} & \underline{24.1} & \underline{21.2} & 18.3 & 12.8 & 8.3 & 4.9 & 2.7 \\
RoPE-Mixed \cite{heoRotaryPositionEmbedding2024} & 28.2 & 24.3 & 20.2 & 16.4 & 12.4 & 6.3 & 2.6 & 1.3 & 0.8 \\
nD-ALiBi \cite{fullerCROMARemoteSensing2023} & 27.2 & 24.5 & 21.2 & 18.3 & 16.2 & 12.8 & 10.2 & 8.1 & 6.2 \\
LookHere \cite{fullerLookHereVisionTransformers2024} & 28.6 & 26.4 & 23.4 & 21.1 & \underline{18.8} & \underline{15.3} & \underline{12.1} & \underline{9.5} & \underline{7.5} \\
\midrule
\textbf{PaPE (ours)} & \textbf{29.4} & \textbf{28.0} & \textbf{26.7} & \textbf{25.4} & \textbf{24.3} & \textbf{22.0} & \textbf{19.7} & \textbf{17.8} & \textbf{15.7} \\
\textbf{PaPE-RI (ours)} & 26.8 & 22.9 & 20.3 & 18.3 & 16.7 & 13.0 & 9.9 & 7.8 & 6.1 \\
        \bottomrule
    \end{tabular}
    }
    \label{tab:extrapolation_renditions}
\end{table*}
\begin{table*}
    \centering
    \caption{
    \textbf{Extrapolation on ObjectNet.}
    PaPE maintains its rank as the \textbf{best} positional encoding at extrapolating, increasing the accuracy at $1024^2$ by 7.8 over the \underline{next-best} (LookHere).
    }
    \resizebox{\linewidth}{!}{
    \begin{tabular}{l|rrrrrrrrr}
        \toprule
        \textbf{Positional Encoding} & $224^2$ & $320^2$ & $384^2$ & $448^2$ & $512^2$ & $640^2$ & $768^2$ & $896^2$ & $1024^2$ \\
        \midrule
nD-sincos \cite{wangTranslatingMathFormula2021} & 28.9 & 24.6 & 20.9 & 15.1 & 10.7 & 5.9 & 3.5 & 2.0 & 1.0 \\
RoPE \cite{suRoFormerEnhancedTransformer2024} & \underline{31.0} & \underline{33.2} & \underline{33.2} & \underline{32.2} & \underline{30.0} & 24.3 & 18.7 & 13.4 & 9.2 \\
RoPE-Mixed \cite{heoRotaryPositionEmbedding2024} & 28.5 & 29.8 & 28.2 & 25.6 & 22.0 & 13.8 & 7.6 & 4.5 & 2.9 \\
nD-ALiBi \cite{fullerCROMARemoteSensing2023} & 29.2 & 31.6 & 30.2 & 28.9 & 26.9 & 23.0 & 20.0 & 16.9 & 14.9 \\
LookHere \cite{fullerLookHereVisionTransformers2024} & 29.8 & 32.3 & 32.0 & 31.2 & 29.9 & \underline{27.1} & \underline{24.4} & \underline{21.6} & \underline{19.1} \\
\midrule
\textbf{PaPE (ours)} & \textbf{31.1} & \textbf{34.3} & \textbf{34.8} & \textbf{34.6} & \textbf{34.1} & \textbf{32.6} & \textbf{30.9} & \textbf{28.7} & \textbf{26.9} \\
\textbf{PaPE-RI (ours)} & 29.4 & 30.5 & 30.3 & 28.7 & 27.0 & 23.3 & 18.7 & 15.7 & 13.2 \\
        \bottomrule
    \end{tabular}
    }
    \label{tab:extrapolation_objectnet}
\end{table*}
We report extrapolation results for the out-of-distribution datasets in \cref{tab:extrapolation_imagenetv2} (ImageNetV2), \cref{tab:extrapolation_renditions} (ImageNet-Renditions), and \cref{tab:extrapolation_objectnet} (ObjectNet).
PaPE consistently ranks the highest across all 3 datasets, with notable increases in accuracy from 7.8 to 9.9 at resolution $1024^2$ over the next-best positional encoding, LookHere.
We find that the extrapolation capability of PaPE is unaffected by the distribution shifts.

\subsection{Parameter-matched}
\label{app:parameter-matched-details}
Here, we describe the experimental setup of the experiments in \cref{sec:parameter_matched}.
We retrain the baselines and PaPE-RI on UCF101, DvsGesture, and ImageNet-1K to match the parameters of PaPE.
We choose UCF101 because it is representative of a significant result and because the 3-fold cross-validation allows for a statistical significance test.
We choose DvsGesture because the results are close (nD-sincos and PaPE are tied as highest scoring), making it representative of the results where the score differences are small.
Additionally, DvsGesture is spatially sparse and temporally dense while UCF101 is the opposite, making the two datasets representative of different types of domains.
We include ImageNet-1K because it is a large-scale dataset, and it is essential for the extrapolation experiments.
Note that PaPE-RI has fewer parameters than PaPE because $m=p$ for PaPE-RI ($p=3$ for UCF101 and DvsGesture and $p=2$ for ImageNet-1K), which is why we also retrain PaPE-RI with matched parameter counts.
PaPE solely adds parameters that are used in the queries and keys.
As such, we consider it the fairest comparison to match parameters by increasing the query and key head sizes of all other methods.
Note also that this setting is particularly favorable to RoPE where larger query-key head sizes not only add more semantic capability, but also increase the capacity of the positional encoding \cite{barberoWeGoWhat2024}.
We do not change the head size of the values as PaPE does not change this either.
A ViT-B on UCF101 has 86.3M parameters while a ViT-B with PaPE ($m=50$) has 97.4M parameters.
The other methods match the additional 11.1M parameters of PaPE by increasing the query-key head size from 64 to 114.
\begin{table*}
    \centering
    \caption{
    \textbf{Parameter-matched per-fold accuracy on UCF101.}
    PaPE (\textbf{best}) confirms its position as highest scoring over RoPE (\underline{next-best}).
    The difference is statistically significant at p=0.026.
    }
    \begin{tabular}{l|ccc|c}
        \toprule
        \textbf{Positional Encoding} & \textbf{Fold 1} & \textbf{Fold 2} & \textbf{Fold 3} & \textbf{Mean ± std} \\
        \midrule
        nD-sincos \cite{wangTranslatingMathFormula2021} & 39.3 & 41.1 & 38.1 & 39.5 ± 1.5 \\
        RoPE \cite{suRoFormerEnhancedTransformer2024} & \underline{45.4} & \underline{46.2} & \underline{44.2} & \underline{45.2} ± 1.0 \\
        RoPE-Mixed \cite{heoRotaryPositionEmbedding2024} & 40.4 & 41.6 & 41.9 & 41.3 ± 0.8 \\
        nD-ALiBi \cite{fullerCROMARemoteSensing2023} & 41.4 & 43.9 & 40.3 & 41.9 ± 1.8 \\
        \midrule
        \textbf{PaPE (ours)} & \textbf{51.4} & \textbf{48.2} & \textbf{49.1} & \textbf{49.5} ± 1.6 \\
        \textbf{PaPE-RI (ours)} & 40.2 & 41.9 & 43.4 & 41.8 ± 1.6 \\
        \bottomrule
    \end{tabular}
    \label{tab:ucf101_crossfold_param_matched}
\end{table*}

We report per-fold accuracies on UCF101 for the parameter-matched experiments in \cref{tab:ucf101_crossfold_param_matched}.
We observe that PaPE stays best with RoPE being next-best, and that the difference remains statistically significant at p=0.026.
\begin{table*}
    \centering
    \caption{
    \textbf{Parameter-matched extrapolation on ImageNet-1K.}
    PaPE preserves its strong classification extrapolation as the \textbf{best} positional encoding with a 9.6 improvement in accuracy at $1024^2$ over the \underline{next-best} (LookHere).
    }
    \resizebox{\linewidth}{!}{
    \begin{tabular}{l|rrrrrrrrr}
        \toprule
        \textbf{Positional Encoding} & $224^2$ & $320^2$ & $384^2$ & $448^2$ & $512^2$ & $640^2$ & $768^2$ & $896^2$ & $1024^2$ \\
        \midrule
nD-sincos \cite{wangTranslatingMathFormula2021} & 79.6 & 76.8 & 72.7 & 65.8 & 58.0 & 39.5 & 22.3 & 11.8 & 6.9 \\
RoPE \cite{suRoFormerEnhancedTransformer2024} & 80.2 & 80.8 & 80.1 & 79.0 & 77.3 & 72.4 & 64.3 & 52.9 & 39.8 \\
RoPE-Mixed \cite{heoRotaryPositionEmbedding2024} & 80.1 & 80.2 & 78.3 & 74.8 & 69.9 & 55.2 & 36.9 & 22.9 & 14.7 \\
nD-ALiBi \cite{fullerCROMARemoteSensing2023} & 80.0 & 80.4 & 79.9 & 79.0 & 77.9 & 75.2 & 71.2 & 65.6 & 59.1 \\
LookHere \cite{fullerLookHereVisionTransformers2024} & \underline{80.4} & \underline{81.5} & \underline{81.3} & \underline{80.8} & \underline{80.1} & \underline{78.2} & \underline{75.2} & \underline{70.6} & \underline{65.3} \\
\midrule
\textbf{PaPE (ours)} & \textbf{80.6} & \textbf{81.6} & \textbf{81.7} & \textbf{81.6} & \textbf{81.3} & \textbf{80.5} & \textbf{79.2} & \textbf{77.3} & \textbf{74.9} \\
\textbf{PaPE-RI (ours)} & 79.8 & 80.3 & 78.2 & 75.7 & 73.2 & 67.7 & 60.9 & 53.4 & 45.7 \\
        \bottomrule
    \end{tabular}
    }
    \label{tab:extrapolation_param_matched}
\end{table*}

\cref{tab:extrapolation_param_matched} reports extrapolation results where the baselines and PaPE-RI are parameter-matched with PaPE.
Matching parameters does not change that PaPE extrapolates noticeably better.
The largest baseline increase in extrapolation is for LookHere (64.4 vs. 65.3 at $1024^2$).
Even with this increase, PaPE still increases the absolute extrapolation accuracy by up to 9.6, showing that its strong classification extrapolation stems from an improved positional encoding.
\begin{table*}[t]
    \centering
    \caption{\textbf{Parameter-matched by reducing hidden size.}
    Reducing the hidden size from $d=768$ to $d=680$ does not diminish the performance of PaPE, supporting that PaPE's advantage stems from an improved positional encoding and not from increasing the number of parameters.}
    \begin{tabular}{l|ccccc}
        \toprule
        \textbf{Size} & \textbf{UCF101} & \textbf{DvsGesture} & \textbf{GEN1} & \textbf{ImageNet-1K} & \textbf{COCO} \\
        \midrule
        Reduced ($d=680)$ & 50.8 & 93.8 & 34.9 & 80.7 & 38.8 \\
        Full ($d=768$) & 49.5 & 93.4 & 34.8 & 80.6 & 38.9 \\
        \bottomrule
    \end{tabular}
    \label{tab:param-matched-reduced}
\end{table*}

Finally, we reduce the hidden size, $d$, of PaPE to match the total parameter count to that of the baselines.
We achieve this by reducing $d$ from 768 to 680.
\cref{tab:param-matched-reduced} reports the results of this experiment.
We find that the reduced version of PaPE does not obtain worse scores than the full version, strengthening the case for PaPE's generality stemming from an improved positional encoding rather than an increase in parameters.

\subsection{UCF101 Per-Fold Accuracy}
\label{app:ucf101_crossfold}
\begin{table*}
    \centering
    \caption{
    \textbf{UCF101 per-fold accuracy.}
    PaPE (\textbf{best}) achieves noticeable gains over RoPE (\underline{next-best}) on all folds.
    The difference is statistically significant at p=0.011.
    }
    \begin{tabular}{l|ccc|c}
        \toprule
        \textbf{Positional Encoding} & \textbf{Fold 1} & \textbf{Fold 2} & \textbf{Fold 3} & \textbf{Mean ± std} \\
        \midrule
        nD-sincos \cite{wangTranslatingMathFormula2021} & 37.8 & 39.1 & 39.3 & 38.7 ± 0.8 \\
        RoPE \cite{suRoFormerEnhancedTransformer2024} & \underline{45.3} & \underline{43.8} & \underline{42.8} & \underline{43.9} ± 1.2 \\
        RoPE-Mixed \cite{heoRotaryPositionEmbedding2024} & 40.6 & 41.1 & 41.9 & 41.2 ± 0.7 \\
        nD-ALiBi \cite{fullerCROMARemoteSensing2023} & 41.2 & 41.5 & 42.1 & 41.6 ± 0.5 \\
        \midrule
        \textbf{PaPE (ours)} & \textbf{51.4} & \textbf{48.2} & \textbf{49.1} & \textbf{49.5} ± 1.6 \\
        \textbf{PaPE-RI (ours)} & 43.0 & 40.2 & 43.3 & 42.2 ± 1.7 \\
        \bottomrule
    \end{tabular}
    \label{tab:ucf101_crossfold}
\end{table*}
We have evaluated UCF101 using the official 3-fold cross-validation splits, but we only reported the mean accuracy in \cref{tab:generality}.
Here, we report the accuracy of each fold as well as the mean and standard deviations in \cref{tab:ucf101_crossfold}.
We find from the additional results that the difference between PaPE (best) and RoPE (next-best) is statistically significant at p=0.011, confirming PaPE's spatio-temporal relevance.

\subsection{Ablation Study: Extrapolation of m}
\label{app:m_extrapolation}
\begin{table*}
    \centering
    \caption{
    \textbf{Ablation of m with extrapolation.}
    We follow the ablation setup in \cref{sec:ablations}.
    $m$ does not have a large effect on extrapolation.
    }
    \begin{tabular}{l|rrrrrrrrr}
        \toprule
        \textbf{m} & $224^2$ & $320^2$ & $384^2$ & $448^2$ & $512^2$ & $640^2$ & $768^2$ & $896^2$ & $1024^2$ \\
        \midrule
        2 & 78.7 & 79.9 & 79.6 & 79.1 & 78.5 & 77.1 & 75.2 & 72.4 & 69.0 \\
        4 & 78.4 & 79.5 & 79.3 & 78.7 & 78.0 & 76.5 & 74.4 & 71.6 & 68.2 \\
        8 & 79.0 & 79.7 & 79.4 & 78.9 & 78.1 & 76.4 & 74.3 & 71.6 & 68.2 \\
        16 & 79.1 & 80.1 & 80.0 & 79.5 & 78.8 & 77.4 & 75.3 & 72.8 & 69.8 \\
        32 & 78.9 & 80.1 & 79.8 & 79.4 & 78.7 & 77.4 & 75.5 & 72.8 & 69.7 \\
        50 & 79.2 & 80.3 & 80.1 & 79.5 & 78.9 & 77.5 & 75.4 & 72.9 & 69.8 \\
        64 & 79.0 & 80.1 & 79.9 & 79.4 & 78.6 & 77.1 & 75.1 & 72.5 & 69.0 \\
        \bottomrule
    \end{tabular}
    \label{tab:extrapolation_m}
\end{table*}
\cref{tab:extrapolation_m} reports extrapolation results for the ablations of $m$ in \cref{sec:ablations}.
The accuracy at $1024^2$ is lowest at 68.2 for $m \in \{ 4, 8 \}$ and highest at 69.8 for $m=50$.
We conclude that $m$ does not affect extrapolation greatly.
Together with the efficiency results reported in \cref{tab:efficiency}, this shows that it is possible to significantly reduce the overhead of PaPE without compromising extrapolation.

\section{Efficiency}
\label{app:efficiency}
\begin{table*}
\centering
\caption{
\textbf{Efficiency.}
Resource usage is averaged over 1000 samples with a batch size of 1, using $224^2$ resolution images from ImageNet-1K.
PaPE with $m=50$ produces a moderate resource increase from 7\% to 20\%.
}
\resizebox{\linewidth}{!}{
\begin{tabular}{lrrrrrrrr}
\toprule
& & & \multicolumn{3}{c}{\textbf{Training}} & \multicolumn{3}{c}{\textbf{Inference}} \\
\cmidrule(lr){4-6} \cmidrule(lr){7-9}
\textbf{Method} & \textbf{Total Params.} & \textbf{Pos. Params.} & \textbf{VRAM (MB)} & \textbf{GFLOPS} & \textbf{Time (ms)} & \textbf{VRAM (MB)} & \textbf{GFLOPS} & \textbf{Time (ms)} \\
\midrule
nD-sincos   & 86.4M & 0     & 1669 & 105.9 & 12.5 & 362 & 35.1 & 1.4 \\
nD-ALiBi    & 86.4M & 0     & 1673 & 105.2 & 12.5 & 365 & 35.1 & 1.4 \\
RoPE        & 86.4M & 0     & 1669 & 105.9 & 12.6 & 363 & 35.1 & 1.4 \\
RoPE-Mixed  & 86.4M & 18.4K & 1671 & 105.9 & 12.7 & 363 & 35.1 & 1.4 \\
LookHere    & 86.4M & 0     & 1694 & 105.2 & 12.6 & 387 & 35.1 & 1.4 \\ \midrule
PaPE (m=2)  & 86.9M & 0.5M  & 1680 & 107.1 & 13.1 & 366 & 35.5 & 1.5 \\
PaPE (m=4)  & 87.3M & 0.9M  & 1688 & 107.6 & 13.1 & 369 & 35.6 & 1.5 \\
PaPE (m=8)  & 88.2M & 1.8M  & 1705 & 108.7 & 13.3 & 374 & 35.7 & 1.5 \\
PaPE (m=16) & 90.0M & 3.6M  & 1778 & 110.8 & 13.5 & 394 & 36.1 & 1.5 \\
PaPE (m=32) & 93.5M & 7.1M  & 1804 & 114.9 & 14.0 & 416 & 36.8 & 1.5 \\
PaPE (m=50) & 97.5M & 11.1M & 1901 & 119.7 & 14.6 & 433 & 37.6 & 1.6 \\
PaPE (m=64) & 100.6M & 14.2M & 1947 & 123.3 & 14.9 & 447 & 38.2 & 1.6 \\
PaPE-RI     & 86.5M & 110.6K & 1674 & 106.4 & 12.9 & 365 & 35.3 & 1.5 \\
\midrule
\makecell{Relative increase\\(Ref. PaPE $m=50$)} & 13\% & - & 14\% & 14\% & 17\% & 20\% & 7\% & 14\% \\
\bottomrule
\end{tabular}
}
\label{tab:efficiency}
\end{table*}
As shown in \cref{sec:efficient_pape}, PaPE increases the effective head size by $p^2 +2p + 2$.
It also introduces positional parameters: $W_a$, $W_b$, and $W_p$.
We measure the number of additional positional parameters and the impact on memory usage, FLOPS, and step times for training as well as inference for different $m$ in \cref{tab:efficiency}.
The analysis considers a ViT-B/16 on ImageNet-1K at resolution $224^2$ and a batch size of 1.

Overall, we find that PaPE increases resource usage moderately by 7\%--20\% for the $m=50$ configuration that is used in the experiments, compared to the lowest resource baseline.
The relative increase is 9\%--23\% for the largest considered option at $m=64$.
This shows that the overhead incurred by PaPE is within a reasonable range, and that PaPE remains a practical positional encoding for vision transformers.

A standard ViT-B (including its classification head) has 86.4M parameters.  
With PaPE, this footprint grows only by 1\%--16\%.
Although this exceeds the baseline sizes, PaPE still delivers its advantages with a moderate increase in parameters.

Step times are measured on an RTX 4090 GPU.
PaPE increases training step time by 0.6--2.4 ms and inference time by 0.1--0.2 ms over the fastest baselines.
In absolute terms, this overhead is tiny, so PaPE remains fast.
However, the relative increase is between 5\%--19\% for training and 7\%--14\% for inference, which may matter for some practical applications.

\section[Motivation for W\_p]{Motivation for $W_p$}
\label{app:wp_m_motivation}
We introduced $W_p \in \mathbb{R}^{m \times p}$ in \cref{eqn:delta_pos} and noted that it is empirically beneficial, even if not strictly required to satisfy our goal.
Here we elaborate on the motivation for introducing it.

$W_p$ is intended to make optimization easier rather than to add model capacity (for $m > \min (d, p)$).
Once $m > \min (d, p)$, PaPE has already spanned the representational space it needs and larger $m$ mainly gives the optimizer more directions to discover useful patterns, rather than attaining greater capacity.

We can view $W_p$ as expressing the $p$ positional dimensions through $m$ learned basis functions.
As such, $W_p$ can ease training by providing more opportunities for exploring different directions than the ones defined by the unprojected $p$ dimensions.
Expanding on this view, we can consider $m$ as a bottleneck below and above a threshold of $\min (d, p)$.

Below the threshold ($m < \min (d, p)$): increasing $m$ should improve both representational capacity and the optimization landscape, since $m$ is limiting what the model can express.

Above the threshold ($m > \min (d, p)$): in the $b$-term, larger $m$ can still ease optimization, but doesn't add representational capacity.
After training, results are unaffected if $W_b$ and $W_p$ are compressed to a full-rank representation.
In the $a$-term the softplus and quadratic nonlinearities mean larger $m$ can still help both representation and optimization, though gains are likely bounded since the $m$ values are aggregated via a dot product.

\section{Model Analysis}
\label{sec:model_analysis}
\begin{figure}
    \centering
    \includegraphics[width=0.5\linewidth]{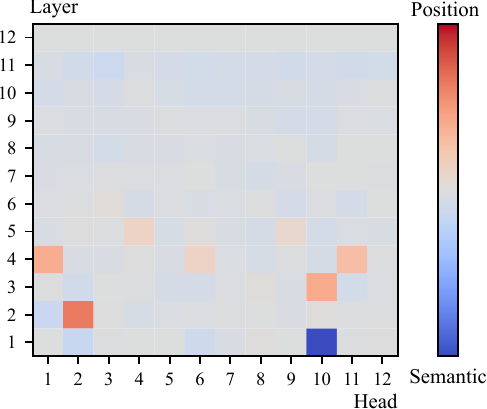}
    \caption{\textbf{Model analysis on ImageNet-1K.}
    Red ($z > 0$) highlights heads that lean heavily on positional information, while blue ($z < 0$) marks heads that prioritize semantic content in deciding what to attend to.
    Positions are used most strongly in early layers.}
    \label{fig:model_analysis}
\end{figure}

The decomposed attention in PaPE eases the investigation of how the model uses position and semantic information.
Although model analysis is not the focus of this work, here we demonstrate an analysis method that is enabled by PaPE.

The attention matrix can---disregarding the dot product scaling factor---be represented as $A_{ij} = \frac{1}{\gamma_i}P_{ij}Y_{ij}$, with positional components 
\begin{align}
    P_{ij}=\exp \left( \left< a_i, \Delta r_{ij}^{\odot 2} \right> + \left< b_i, \Delta r_{ij} \right> \right),    
\end{align}
semantic components $Y_{ij} = \exp \left( \left< q_i, k_j \right> \right)$, and normalization factors $\gamma_i = \sum_j P_{ij}Y_{ij}$.
We introduce the score $z$,
\begin{align}
    z = \mathbb{E}_{ij} \left[ \tfrac{1}{\gamma_i}(P_{ij} - Y_{ij}) \right], \label{eq:scores-model-analysis}
\end{align}
to quantify the relative importance of positional vs. semantic terms of a given attention head.
Since attention patterns are often sparse, it is helpful to refine the analysis by considering only the top-attended keys for each query, which we define as the minimal set of keys whose cumulative attention in $A_i$ reaches a threshold $\tau$.

\cref{fig:model_analysis} breaks down how each head in the ImageNet-1K model focuses on positions and semantics on average by visualizing the scores $z$, using the threshold $\tau=80\%$.
Most heads balance positions and semantics quite evenly.
Yet, a few stand out: some heads (e.g., L2H2) are clearly position-focused, whereas L1H10 is highly driven by semantics.
Interestingly, these specialized heads cluster in the early layers (L1–L5), suggesting that emphasizing positional or semantic signals is especially beneficial when transforming low-level inputs into higher-level representations.

\section{Comparison of nD-ALiBi and PaPE-RI}
\label{sec:alibi-and-pape}
The methods nD-ALiBi and PaPE-RI have methodological similarities, which we compare here.

nD-ALiBi works by subtracting the distance matrix $D$,
\begin{align}
    D_{ij} = \lVert r_j - r_i\rVert_2,
\end{align}
between token positions from the $S$ matrix (pre-softmax attention matrix), with a constant scaling factor for each attention head.
For simplicity, we consider the case of one head without loss of generality,
\begin{align}
    S_{\text{ALiBi}}=QK^T + \alpha D \\ 
    \alpha \in \mathbb{R}_{<0}.
\end{align}
Similar behavior is obtained by PaPE-RI, inducing the same properties---translation and rotation invariance.
The main difference is that PaPE-RI is designed to subtract the \textit{squared} distance matrix, where the coefficient can also depend on the token content, leading to
\begin{align}
    S_\text{PaPE-RI} = QK^T + \alpha(X) \odot D^2\\
    \alpha_{ij}(X) = \alpha_{ij}(x_i) \in \mathbb{R}_{<0}.
\end{align}
A further difference is that PaPE-RI is compatible with efficient attention kernels, while nD-ALiBi is not.  

\section{Training Details}
\label{app:training_details}
We now detail the training configurations used for all datasets.
All models are optimized with AdamW \cite{loshchilovDecoupledWeightDecay2019}, using a OneCycle \cite{smithSuperconvergenceVeryFast2019} learning rate schedule with cosine decay.
The only exception is UniTR on nuScenes, which uses Adam \cite{kingmaAdamMethodStochastic2015}.
\cref{tab:hyperparameters} summarizes the common hyperparameters for all datasets.
Unless otherwise specified, all models use the same Transformer settings (i.e. \#layers, \#heads, dimensionality, and head size) as ViT-B \cite{dosovitskiyImageWorth16x162020}.
Dataset-specific configurations are detailed in the following paragraphs.

\begin{table*}
    \centering
    \caption{
    \textbf{Hyperparameters.}
    The hyperparameters are shared for all positional encodings for a given dataset.
    }
    \resizebox{\linewidth}{!}{
    \begin{tabular}{c|cccccccc}
    \toprule
        \textbf{Parameter} & \textbf{UCF101} & \textbf{DvsGesture} & \textbf{GEN1} & \textbf{ImageNet-1K} & \textbf{COCO} & \textbf{ScanNet} & \textbf{ModelNet40} & \textbf{nuScenes} \\
        \midrule
        Batch size & 32 & 32 & 64 & 1024 & 64 & 12 & 32 & 3 \\
        Learning rate & 0.00003 & 0.00005 & 0.0001 & 0.0006 & 0.0006 & 0.006 & 0.001 & 0.003 \\
        Weight decay & 0.05 & 0.05 & 0.05 & 0.05 & 0.05 & 0.05 & 0.01 & 0.03 \\
        Momentum ($B_1)$ & 0.9 & 0.9 & 0.9 & 0.9 & 0.9 & 0.9 & 0.9 & 0.9 \\
        Momentum ($B_2)$ & 0.999 & 0.999 & 0.999 & 0.999 & 0.999 & 0.999 & 0.999 & 0.99 \\
        Warmup epochs & 20 & 15 & 7.5 & 15 & 15 & 40 & 5 & 1 \\
        Total epochs & 200 & 300 & 150 & 300 & 150 & 800 & 300 & 10 \\
        Training resolution ($W \times H$) & $320 \times 240$ & $128^2$ & $304 \times 240$ & $224^2$ & $640^2$ & - & - & $704 \times 256$ \\
        Patch size & 16 & 16 & 16 & 16 & 16 & 1024 & 1024 & 8 \\
        Drop path & 0.1 & 0.1 & 0.1 & 0.1 & 0.1 & 0.3 & 0.3 & - \\
        Label smoothing & - & 0.1 & - & 0.1 & - & - & - & - \\
        \bottomrule
    \end{tabular}
    }
    \label{tab:hyperparameters}
\end{table*}

\textbf{UCF101.}
We follow ViViT \cite{arnabViViTVideoVision2021}, using 2-tubelets for each video sample and drawing up to 5 samples per video, with 10 frames separating consecutive 2-tubelets.
The augmentations are RandAugment and HorizontalFlip.
For both training and evaluation, we use the official 3-fold cross-validation splits and report the mean test performance.
Mirroring the ImageNet-1K protocol, we further partition each training split into new training and validation sets.
We do this by assigning the first video group of every action class to the validation set.

\textbf{DvsGesture.}
We utilize Spiking Patches \cite{ohrstrom2025spiking} to extract asynchronous and spatially sparse spatio-temporal tokens, configured with a spike threshold, $\sigma$, of 256 and a refractory period, $T$, of 100 ms.
The model is a ViT-B where we follow the token embedding method of \citet{ohrstrom2025spiking}.
Since the temporal positions are at a microsecond resolution, we rescale them by $1/50000$.
We apply two sets of augmentations: NDA \cite{liNeuromorphicDataAugmentation2022} and EventDrop \cite{guEventDropDataAugmentation2021}.

\textbf{GEN1.}
Similar to DvsGesture, we extract tokens using Spiking Patches with $\sigma = 256$ and $T = 100$ ms, and process them with a ViT-B backbone.  
Here, however, we scale temporal positions by $1/100000$ and sample the most recent 500K events relative to the prediction time.
Because the tokens are asynchronous, they are not directly compatible with the image-like tensors expected by the ViTDet neck and YOLOv10 head.
Therefore, we first convert the backbone outputs into an image-like tensor.
The way it works is that for each spatial location, we fill its slot with the most recent token at that spatial position.
We augment with EventDrop and rotations.

\textbf{ImageNet-1K.}
We treat the official validation split as our test set and divide the original training split into a new training set and a validation set.
The ratio is 99\% for training and 1\% for validation, using a stratified split that preserves class balance in each subset.
For the classifier head, we initialize the weights to 0 and the bias to $-\log(1000)$, so that the initial class probabilities are uniform.
We use the following set of augmentations:
RandAugment \cite{cubukRandaugmentPracticalAutomated2020}, HorizontalFlip, MixUp \cite{zhangMixupEmpiricalRisk2018}, and CutMix \cite{yunCutMixRegularizationStrategy2019}.

\textbf{COCO.}
The detection model pairs a ViT-B backbone with a ViTDet \cite{liExploringPlainVision2022} neck (strides 8, 16, and 32) and a YOLOv10 \cite{wangYOLOv10RealTimeEndtoEnd2024a} head and loss.
For data augmentation, we keep RandAugment but explicitly drop shear and rotation, as they corrupt ground-truth bounding boxes.
In addition, we apply Large Scale Jitter \cite{ghiasi2020simple}, following ViTDet.

\textbf{ScanNet and ModelNet40.}
We use Point Transformer V3 \cite{wuPointTransformerV32024} with the dataset-specific configurations from the official code repository.
This includes the same augmentations as well as model configuration and size.
However, we do not initialize the weights from pretrained models.

\textbf{nuScenes.}
We use UniTR \cite{wangUniTRUnifiedEfficient2023}, following the same architecture, model size, hyperparameters, and augmentations as in the official code repository.

\end{document}